\documentclass[runningheads]{llncs}
\usepackage{graphicx}
\usepackage{comment}
\usepackage{amsmath,amssymb}
\usepackage{color}
\usepackage{url}
\usepackage{tabulary}
\usepackage{booktabs}
\usepackage{amsfonts}
\usepackage{nicefrac}
\usepackage{microtype}
\usepackage{algorithm}
\usepackage{algorithmic}
\usepackage{bm}
\usepackage{graphicx}
\usepackage{subfigure} 
\usepackage{multirow}
\usepackage{comment}
\usepackage{float}
\usepackage{indentfirst}
\usepackage{caption}
\usepackage{hyperref}
\usepackage[misc]{ifsym}

\usepackage[dvipsnames]{xcolor}
\hypersetup{
	colorlinks   = true,
	linkcolor    = NavyBlue,
	citecolor   = NavyBlue
}

\begin{document}

\pagestyle{headings}
\mainmatter
\def\ECCVSubNumber{3769}

\title{Minimum Class Confusion for Versatile Domain Adaptation}

\begin{comment}
\titlerunning{ECCV-20 submission ID \ECCVSubNumber} 
\authorrunning{ECCV-20 submission ID \ECCVSubNumber} 
\author{Anonymous ECCV submission}
\institute{Paper ID \ECCVSubNumber}
\end{comment}

\titlerunning{Minimum Class Confusion for Versatile Domain Adaptation}

\author{
   Ying Jin \and
   Ximei Wang \and
   Mingsheng Long~(\Letter) \and
   Jianmin Wang}
\authorrunning{Jin et al.}

\institute{
   School of Software, BNRist, Research Center for Big Data, Tsinghua University, China\\
   %Research Center for Big Data, Tsinghua University, China\\
\email{\{jiny18,wxm17\}@mails.tsinghua.edu.cn, \{mingsheng,jimwang\}@tsinghua.edu.cn}}

\maketitle

\begin{abstract}
There are a variety of Domain Adaptation (DA) scenarios subject to label sets and domain configurations, including closed-set and partial-set DA, as well as multi-source and multi-target DA. It is notable that existing DA methods are generally designed only for a specific scenario, and may underperform for scenarios they are not tailored to. To this end, this paper studies \textbf{Versatile Domain Adaptation (VDA)}, where one method can handle several different DA scenarios without any modification. Towards this goal, a more general inductive bias other than the \emph{domain alignment} should be explored. We delve into a missing piece of existing methods: \emph{class confusion}, the tendency that a classifier confuses the predictions between the correct and ambiguous classes for target examples, which is common in different DA scenarios. We uncover that reducing such pairwise class confusion leads to significant transfer gains. With this insight, we propose a general loss function: \textbf{Minimum Class Confusion (MCC)}. It can be characterized as (1) a \emph{non-adversarial} DA method without explicitly deploying domain alignment, enjoying faster convergence speed; (2) a \emph{versatile} approach that can handle four existing scenarios: Closed-Set, Partial-Set, Multi-Source, and Multi-Target DA, outperforming the state-of-the-art methods in these scenarios, especially on one of the largest and hardest datasets to date ($7.3\%$ on DomainNet). Its versatility is further justified by two scenarios proposed in this paper: Multi-Source Partial DA and Multi-Target Partial DA. In addition, it can also be used as a general regularizer that is orthogonal and complementary to a variety of existing DA methods, accelerating convergence and pushing these readily competitive methods to stronger ones. Code is available at \url{https://github.com/thuml/Versatile-Domain-Adaptation}.

\end{abstract}

\section{Introduction}

The scarcity of labeled data hinders deep neural networks (DNNs) from use in real applications. This challenge gives rise to Domain Adaptation (DA)~\cite{Quionero2009Dataset,pan2010survey}, an important technology that aims to transfer knowledge from a labeled source domain to an unlabeled target domain in the presence of dataset shift.
A rich line of DNN-based methods~\cite{Tzeng14DDC,Long15DAN,Long16RTN,Long17JAN,AFN2019,Tzeng15SDT,DANN2016,ADDA2017,Pei18MADA,Long18CDAN,wang2019tada,MDD2019} have been proposed for Unsupervised DA (UDA), a closed-set scenario with one source domain and one target domain sharing the same label set. Recently, several highly practical scenarios were proposed, such as Partial DA (PDA)~\cite{SAN2018,IWAN2018} with the source label set subsuming the target one, Multi-Source DA (MSDA)~\cite{MDAN2018,xu2018deep}  with multiple source domains, and Multi-Target DA (MTDA)~\cite{Peng2019DADA} with multiple target domains. As existing UDA methods cannot be applied directly to these challenging scenarios, plenty of methods~\cite{SAN2018,Cao2017PADA,IWAN2018,xu2018deep,Peng2019DADA} have been designed for each specific scenario, which work quite well in each tailored scenario.

\begin{figure}[pt]
   \begin{center}
      \includegraphics[width=\linewidth]{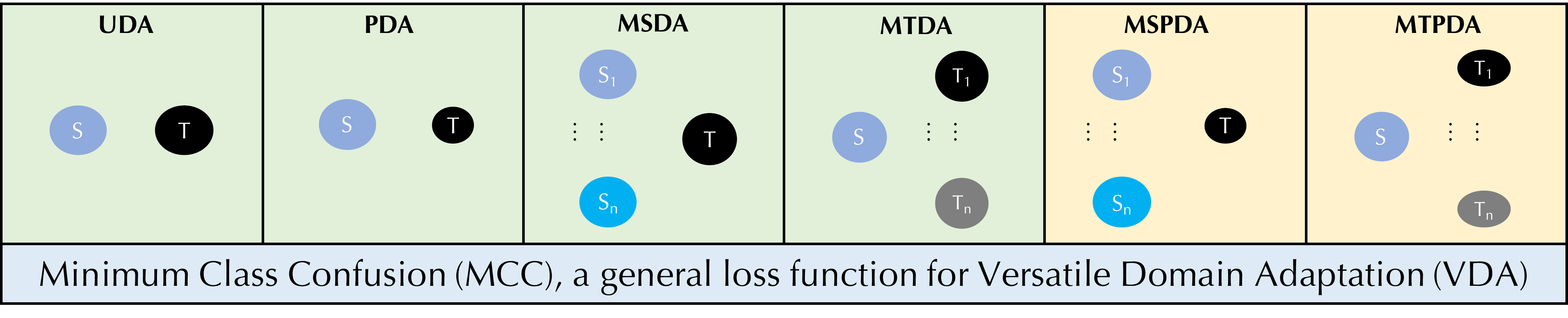}
   \end{center}
   \caption{\textbf{Versatile Domain Adaptation (VDA)} subsumes typical domain adaptation scenarios: (1) Unsupervised Domain Adaptation (UDA); (2) Partial Domain Adaptation (PDA); (3) Multi-Source Domain Adaptation (MSDA); (4) Multi-Target Domain Adaptation (MTDA); (5) Multi-Source Partial Domain Adaptation (MSPDA); (6) Multi-Target Partial Domain Adaptation (MTPDA). Note that scenarios (5)--(6) are newly proposed in this paper. Our \textbf{Minimum Class Confusion (MCC)} is a \emph{versatile} method towards all these DA scenarios.}
   \label{versatile}
\end{figure}

In practical applications, however, it is difficult to confirm the label sets and domain configurations in the data acquisition process. 
Therefore, we may be stuck in choosing a proper method tailored to the suitable DA scenario. 
The most ideal solution to escape from this dilemma is a \emph{versatile} DA method that can handle various scenarios without any modification.
Unfortunately, existing DA methods are generally designed only for a specific scenario and may underperform for scenarios they are not tailored to.  
For instance, PADA~\cite{Cao2017PADA}, a classic PDA method, excels at selecting out outlier classes but suffers from the internal domain shift in MSDA and MTDA, while DADA~\cite{Peng2019DADA}, an outstanding method tailored to MTDA, cannot be directly applied to PDA or MSDA. 
Hence, existing DA methods are not versatile enough to handle practical scenarios of complex variations.

\begin{figure*}[t]
   \centering
   \subfigure[Source]{
      \includegraphics[width=0.205\textwidth]{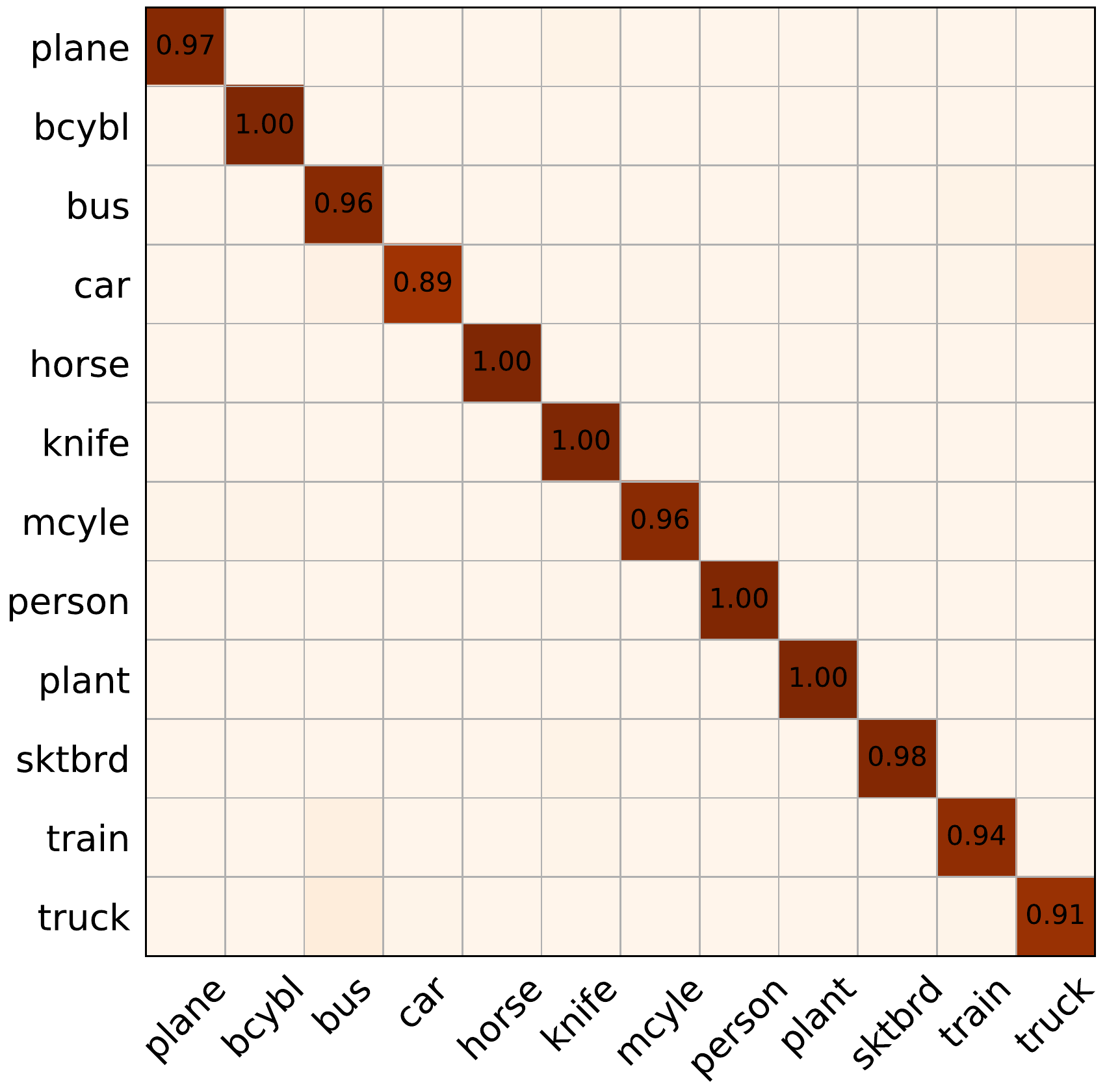}
      \label{ErrorMatrix:source}
   }
   \hfil
   \subfigure[Target]{
      \includegraphics[width=0.24\textwidth]{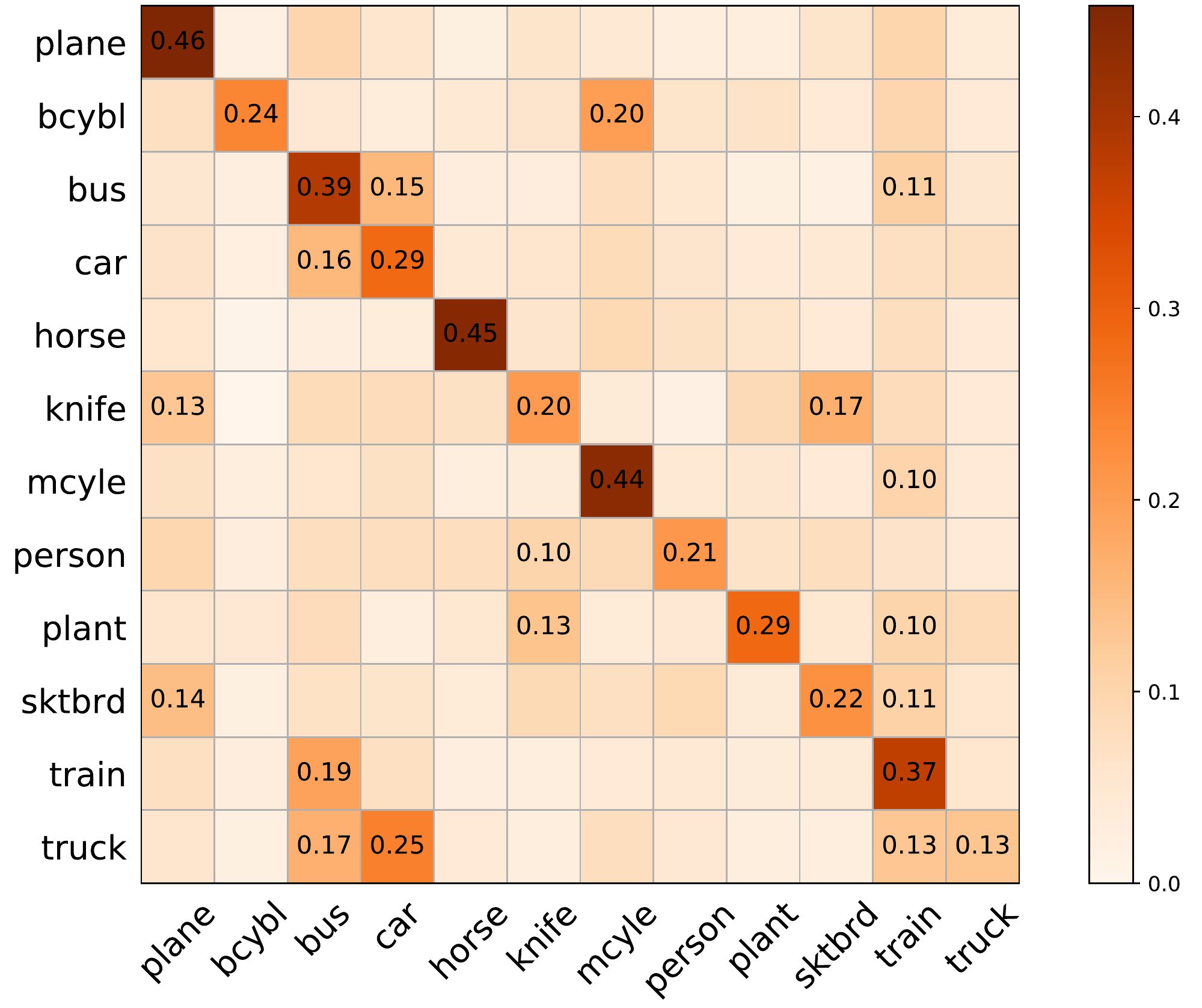}
      \label{ErrorMatrix:target}
   }
   \hfil
   \subfigure[MinEnt]{
      \includegraphics[width=0.205\textwidth]{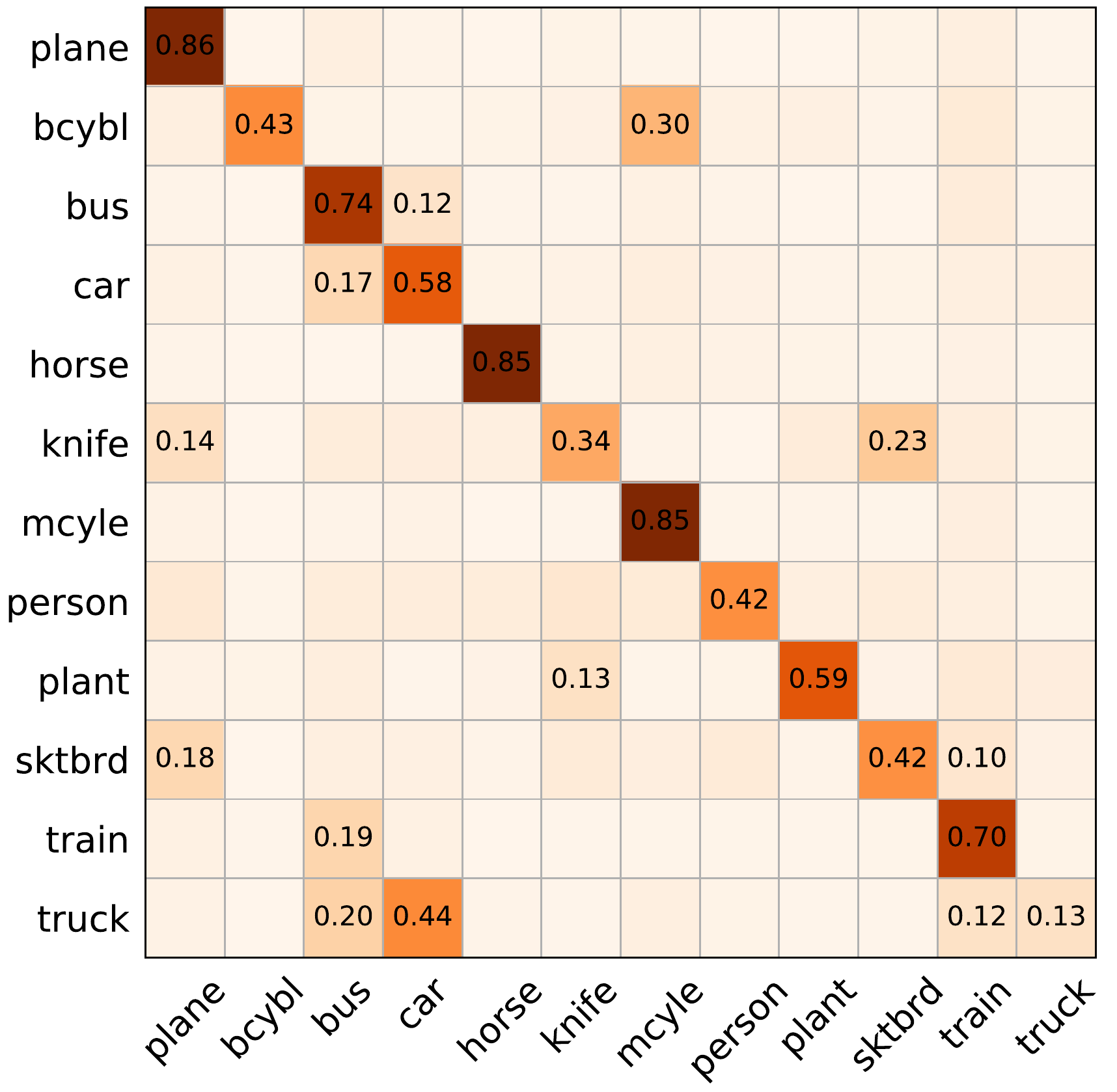}
      \label{ErrorMatrix:MinEnt}
   }
   \hfil
   \subfigure[MCC]{
      \includegraphics[width=0.24\textwidth]{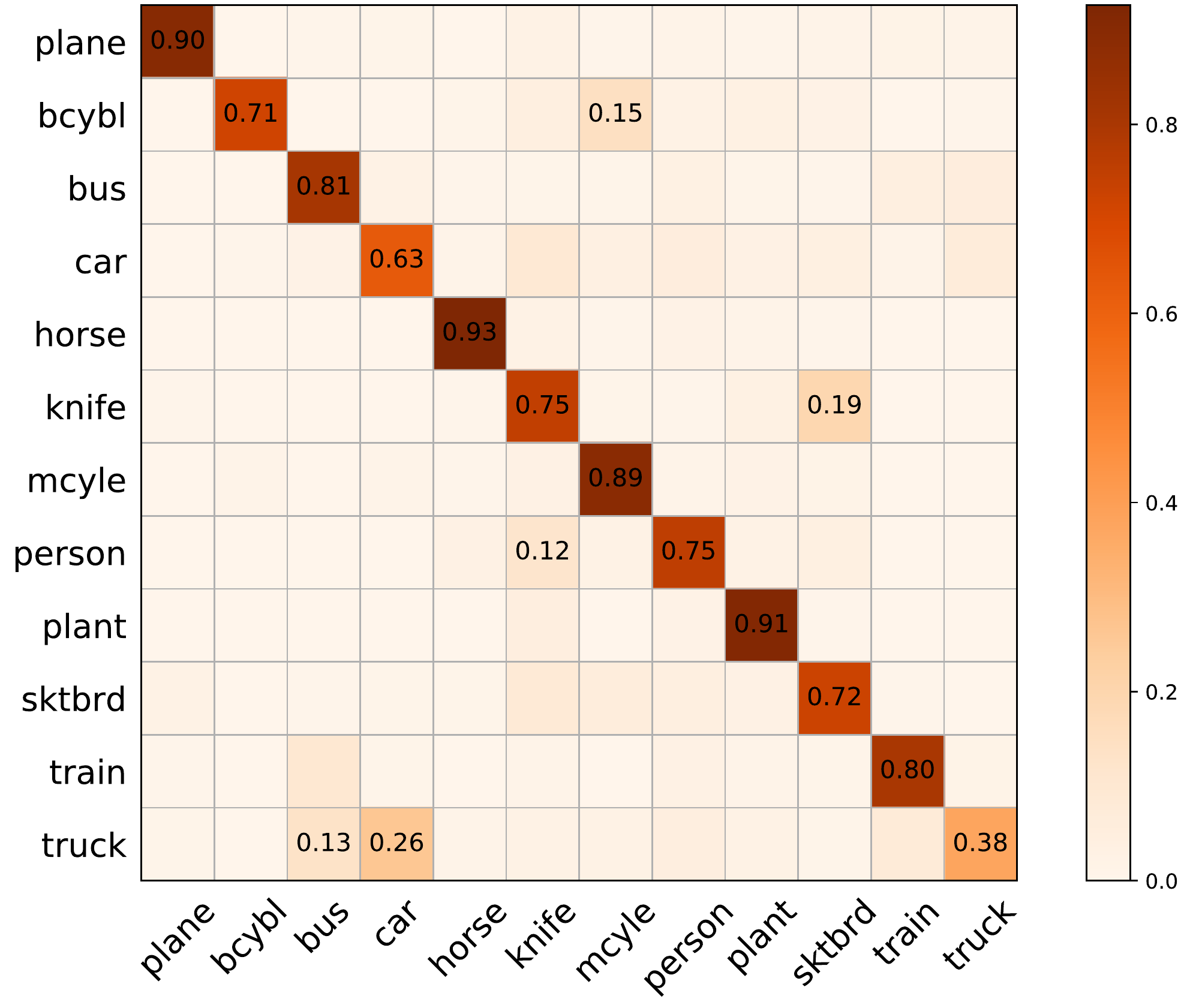}
      \label{ErrorMatrix:MCC}
   }
   \caption{The error matrices of several models on VisDA-2017~\cite{peng2017VisDA}. (a)--(b): Source-only model tested on the source and target domains, showing severe class confusion on target domain examples. (c)--(d): Models trained with entropy minimization (MinEnt) \cite{Grandvalet2005Semi} and Minimum Class Confusion (MCC) on target domain examples, respectively. The proposed MCC loss substantially diminishes the class confusion.}
\end{figure*}

In this paper, we define \textbf{Versatile Domain Adaptation (VDA)} as a line of \emph{versatile} approaches able to tackle a variety of scenarios without any modification. 
Towards VDA, a more general inductive bias other than the domain alignment should be explored.
In this paper, we delved into the error matrices of the target domain and found that the classifier trained on the source domain may confuse to distinguish the correct class from a similar class, such as \texttt{cars} and \texttt{trucks}. 
As shown in Fig.~\ref{ErrorMatrix:target}, 
the probability that a source-only model misclassifies \texttt{cars} as \texttt{trucks} on the target domain is over $25\%$. 
Further, we analyzed the error matrices in other DA scenarios and reached the same conclusion. These findings give us a fresh perspective to enable VDA: \textbf{class confusion}, the tendency that a classifier confuses the predictions between the correct and ambiguous classes for target examples.
We uncover that less class confusion leads to more transfer gains for all the domain adaptation scenarios in Fig.~\ref{versatile}.

Still, we need to address a new challenge that the ground-truth class confusion cannot be calculated if the labels in the target domain are inaccessible. Fortunately, the confusion between different classes can be naturally reflected by an example-weighted inner-product between the classifier predictions and their transposes. And we can define class confusion from this perspective, enabling it to be computed from well-calibrated classifier predictions. To this end, we propose a novel loss function: \textbf{Minimum Class Confusion (MCC)}. It can be characterized as a novel and versatile DA approach without explicitly deploying domain alignment \cite{Long15DAN,DANN2016}, enjoying fast convergence speed. In addition, it can also be used as a general regularizer that is orthogonal and complementary to existing DA methods, further accelerating and improving those readily competitive methods.  Our contributions are summarized as follows:
\begin{itemize}
   \item We propose a practical setting, \textbf{Versatile Domain Adaptation (VDA)}, where one method can tackle many DA scenarios without modification. 
   \item We uncover that the class confusion is a common missing piece of existing DA methods and that less class confusion leads to more transfer gains. 
    \item We propose a novel loss function: \textbf{Minimum Class Confusion (MCC)}, which is versatile to handle four existing DA scenarios, including closed-set, partial-set, multi-source, and multi-target, as well as two proposed scenarios: multi-source partial DA and multi-target partial DA.
   \item We conduct extensive experiments on four standard DA datasets, and show that MCC outperforms the state-of-the-art methods in different DA scenarios, especially on one of the largest and hardest datasets ($\mathbf{7.3\%}$ on DomainNet), enjoying a faster convergence speed than mainstream DA methods. 
\end{itemize}

\section{Related Work}\label{relate}

\textbf{Unsupervised Domain Adaptation (UDA).} 
Most of the existing domain adaptation researches focused on UDA, in which numerous UDA methods were proposed based on either \textit{Moment Matching} or \textit{Adversarial Training}. 

Moment Matching methods aim at minimizing the distribution discrepancy across domains. Deep Coral~\cite{coral2016} aligns second-order statistics between distributions. DDC~\cite{Tzeng14DDC} and DAN~\cite{Long15DAN} utilize Maximum Mean Discrepancy~\cite{MKMMD2012}, JAN~\cite{Long17JAN} defines Joint Maximum Mean Discrepancy, SWD~\cite{SWD2019} introduces Sliced Wasserstein Distance and CAN~\cite{CAN2019} leverages Contrastive Domain Discrepancy. 

Adversarial Training methods were inspired by the Generative Adversarial Networks (GANs)~\cite{Goodfellow14GAN}, aiming at learning domain invariant features in an adversarial manner. DANN~\cite{DANN2016} introduces a domain discriminator to distinguish source and target features, while the feature extractor strives to fool it. ADDA~\cite{ADDA2017}, MADA~\cite{Pei18MADA} and MCD~\cite{MCD2018} extend this architecture to multiple feature extractors and classifiers. Motivated by Conditional GANs~\cite{CGAN2014}, CDAN~\cite{Long18CDAN} aligns domain features in a class-conditional adversarial game. CyCADA~\cite{Hoffman18CyCADA} adapts features in both pixel and feature levels. TADA~\cite{wang2019tada} introduces the first transferable attention mechanism.
SymNet~\cite{symDA2019} uses a symmetric classifier, and 
DTA~\cite{DTA2019} learns discriminative features with a new adversarial dropout. 

There are other approaches to domain adaptation. For instance, SE~\cite{SE2018} is based on the teacher-student~\cite{MeanTea} architecture. TransNorm~\cite{TransNorm} tackles DA with a new transferable backbone.
TAT~\cite{TAT2019} proposes transferable adversarial training to guarantee the adaptability.
BSP~\cite{BSP2019} balances between the transferability and discriminability. 
AFN~\cite{AFN2019} enlarges feature norm to enhance feature transferability. Some methods~\cite{DIRT,ProDA2018,CBST,Zou2019} also utilize the less-reliable self-training or pseudo labeling, {\emph{e.g.} TPN~\cite{TPN2019} is based on pseudo class-prototypes.

\textbf{Partial Domain Adaptation (PDA).}
In PDA, the target label set is a subset of the source label set. Representative methods include SAN~\cite{SAN2018}, IWAN~\cite{IWAN2018}, PADA~\cite{Cao2017PADA} and ETN~\cite{ETN2019}, introducing different weighting mechanisms to select out outlier source classes in the process of domain feature alignment.

\textbf{Multi-Source Domain Adaptation (MSDA).}
In MSDA, there are multiple source domains of different distributions. MDAN~\cite{MDAN2018} provides theoretical insights for MSDA, while Deep Cocktail Network~\cite{xu2018deep} (DCTN) and $\rm M^{3}SDA$~\cite{peng2018moment} extend adversarial training and moment-matching to MSDA, respectively. 

\textbf{Multi-Target Domain Adaptation (MTDA).}
In MTDA, we need to transfer a learning model to multiple unlabeled target domains. DADA~\cite{Peng2019DADA} is the first approach to MTDA through disentangling domain-invariant representations.

\section{Approach}

In this paper, we study \textbf{Versatile Domain Adaptation (VDA)} where one method can tackle many scenarios without any modification. We justify the versatility of one method by four existing scenarios: \textbf{(1)} Unsupervised Domain Adaptation (\textbf{UDA}) \cite{DANN2016}, the standard scenario with a labeled source domain $\mathcal{S} = \{(\mathbf{x}_s^i,{\bf y}_s^i)\}_{i=1}^{n_s}$ and an unlabeled target domain $\mathcal{T} = \{{\mathbf{x}}_t^i\}_{i=1}^{n_t}$, where ${\bf x}^i$ is an example and ${\bf y}^i$ is the associated label; \textbf{(2)} Partial Domain Adaptation (\textbf{PDA}) \cite{Cao2017PADA}, which extends UDA by relaxing the source domain label set to subsume the target domain label set; \textbf{(3)} Multi-Source Domain Adaptation (\textbf{MSDA}) \cite{peng2018moment}, which extends UDA by expanding to $S$ labeled source domains $\{\mathcal{S}_{1}, \mathcal{S}_{2},..., \mathcal{S}_{S}\}$; \textbf{(4)} Multi-Target Domain Adaptation (\textbf{MTDA}) \cite{Peng2019DADA}, which extends UDA by expanding to $T$ unlabeled target domains $\{\mathcal{T}_{1}, \mathcal{T}_{2},..., \mathcal{T}_{T}\}$. We further propose two scenarios to confirm the versatility: \textbf{(5)/(6)} Multi-Source/Multi-Target Partial Domain Adaptation (\textbf{MSPDA/MTPDA}), which extend PDA to multi-source/multi-target scenarios. Tailored to a specific scenario, existing methods fail to readily handle these scenarios. We propose \textbf{Minimum Class Confusion (MCC)} as a generic loss function for VDA. Hereafter, we denote by ${\bf y}_{i\cdot}$, ${\bf y}_{\cdot j}$ and ${\bf Y}_{ij}$ the $i$-row, the $j$-th column and the $ij$-th entry of matrix ${\bf Y}$, respectively.

\begin{figure*}[pt]
   \begin{center}
   \includegraphics[width=1\linewidth]{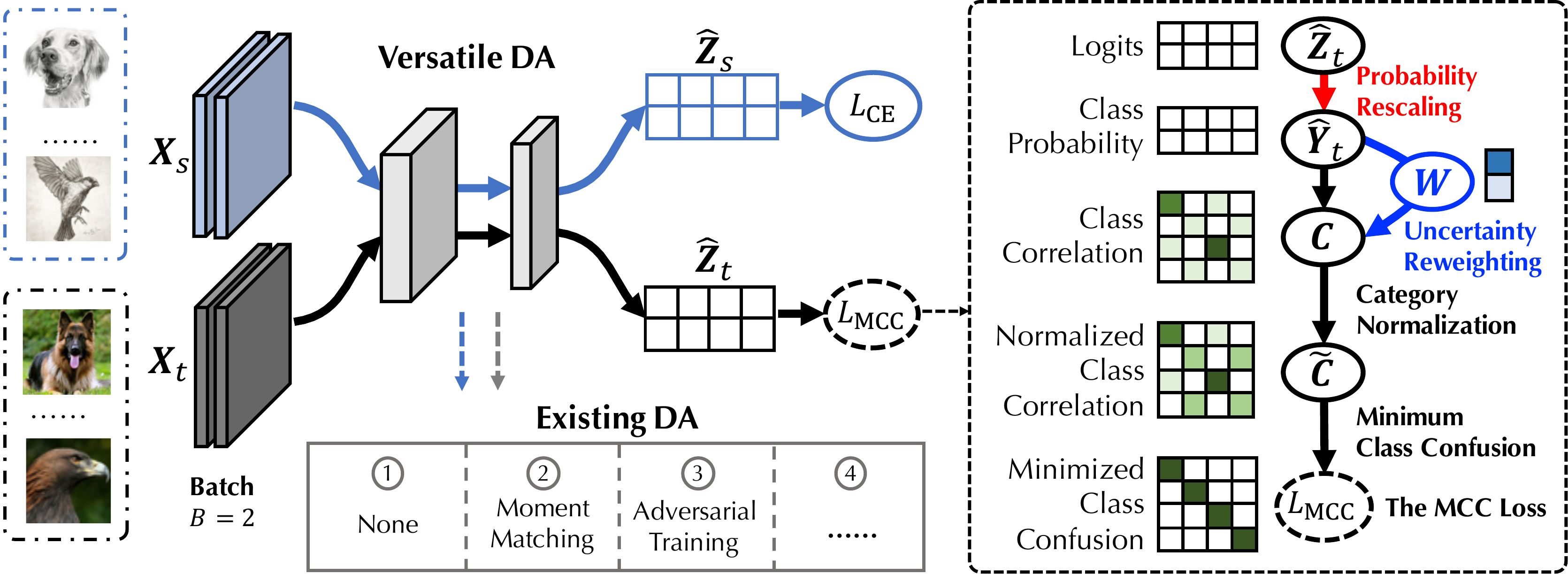}
   \end{center}
   \caption{The schematic of the \textbf{Minimum Class Confusion (MCC)} loss function. Given the shared feature extractor $F$, MCC is defined on the class predictions $\widehat{\mathbf{Y}}_t$ given by the source classifier $G$ on the target data. MCC is versatile to address various domain adaptation scenarios standalone, or to be integrated with existing methods (moment matching, adversarial training, etc). (\textit{Best viewed in color.})}
   \label{method}
\end{figure*}

\subsection{Minimum Class Confusion}

To enable versatile domain adaptation, we need to find out a proper criterion to measure the pairwise class confusion on the target domain. Unlike previous methods such as CORAL~\cite{coral2016} that focus on features, we explore the class predictions. Denote the classifier output on the target domain as ${\widehat{\bf Y}}_t = G(F(\mathbf{X}_t)) \in \mathbb{R }^{B \times |{\cal {C}}|}$, where $B$ is the batch size of the target data, $|{\cal {C}}|$ is the number of source classes, $F$ is the feature extractor and $G$ is the classifier. In our method, we focus on the classifier predictions $\widehat{\bf Y}$ and omit the domain subscript $t$ for clarity. 

\textbf{Probability Rescaling.}
According to \cite{ICML17Calibration}, DNNs tend to make overconfident predictions, hindering them from directly reasoning about the class confusion. Therefore, we adopt temperature rescaling~\cite{hinton2015distilling,ICML17Calibration}, a simple yet effective technique, to alleviate the negative effect of overconfident predictions. Using temperature scaling, the probability $\widehat{Y}_{ij}$ that the $i$-{th} instance belongs to the $j$-{th} class can be recalibrated as 
\begin{equation}\label{softmax}
   \centering
   {\widehat Y_{ij}} = \frac{{\exp \left( {{Z_{ij}}/T} \right)}}{{\sum\nolimits_{j' = 1}^{|{\mathcal{C}}|} {\exp \left( {{Z_{ij'}}/T} \right)} }},
\end{equation} 
where $Z_{ij}$ is the logit output of the classifier layer and $T$ is the temperature hyper-parameter for probability rescaling. Obviously, Eq.~\eqref{softmax} boils down to the vanilla softmax function when $T=1$. 

\textbf{Class Correlation.}
As $\widehat{Y}_{ij}$ reveals the relationship between the $i$-{th} instance and the $j$-{th} class, we define the class correlation between two classes $j$ and $j'$ as
\begin{equation}\label{eq:CC}
   {{\mathbf{C}}_{jj'}} = {\widehat{\mathbf{y}}}_{ \cdot j}^{\sf T}{{\widehat{\mathbf{y}}}_{ \cdot j'}}.
\end{equation}
It is a coarse estimation of the class confusion. Lets delve into the definition of the class correlation in Eq.~\eqref{eq:CC}. Note that ${\widehat{\mathbf{y}}_{ \cdot j}}$ denotes the probabilities that the $B$ examples in each batch come from the $j$-th class. The class correlation measures the possibility that the classifier simultaneously classifies the $B$ examples into the $j$-th and the $j'$-th classes. It is noteworthy that such pairwise class correlation is relatively safe: for false predictions with high confidence, the corresponding class correlation is still low. In other words, highly confident false predictions will negligibly impact the class correlation. 

\textbf{Uncertainty Reweighting.}
We note that examples are not equally important for quantifying class confusion. When the prediction is closer to a uniform distribution, showing no obvious peak (obviously larger probabilities for some classes), we consider the classifier as ignorant of this example. On the contrary, when the prediction shows several peaks, it indicates that the classifier is reluctant between several ambiguous classes (such as \texttt{car} and \texttt{truck}). Obviously, these examples that make the classifier ambiguous across classes will be more suitable for embodying class confusion. As defined in Eq.~\eqref{eq:CC}, these examples can be naturally highlighted with higher probabilities on the several peaks. Further, we introduce a \textit{weighting} mechanism based on uncertainty such that we can quantify class confusion more accurately. Here, those examples with higher certainty in class predictions given by the classifier are more reliable and should contribute more to the pairwise class confusion. We use the \emph{entropy} function $H(p) \triangleq  - {\mathbb{E}_p}\log p$ in information theory as an uncertainty measure of distribution $p$. The entropy (uncertainty) $H(\widehat{\bf y}_{i\cdot})$ of predicting the $i$-{th} example by the classifier is defined as
\begin{equation}
   \centering
   H(\widehat{\bf y}_{i\cdot})= - { \sum _{j=1 }^{ |{\cal {C}}| }{ { \widehat { Y }  }_{ ij }\log{ \widehat { Y }  }_{ ij } }  }.
\end{equation} 
While the entropy is a measure of uncertainty, what we want is a probability distribution that places larger probabilities on examples with larger certainty of class predictions. A \emph{de facto} transformation to probability is the softmax function
\begin{equation}
   {W_{ii}} = \frac{{B\left( {1 + \exp ( { - H( {{{{\widehat{\bf y}}}_{i \cdot }}} )} )} \right)}}{{\sum\limits_{i' = 1}^B {\left( {1 + \exp ( { - H( {{{{\widehat{\bf y}}}_{i' \cdot }}} )} )} \right)} }},
\end{equation}
where $W_{ii}$ is the probability of quantifying the importance of the $i$-th example for modeling the class confusion, and $\mathbf{W}$ is the corresponding diagonal matrix. Note that we take the opposite value of the entropy to reflect the \emph{certainty}. 
Laplace Smoothing~\cite{Laplace2008} (\emph{i.e.} adding a constant $1$ to each addend of the softmax function) is used to form a \emph{heavier-tailed} weight distribution, which is suitable for highlighting more certain examples as well as avoiding overly penalizing the others. For better scaling, the probability over the examples in each batch of size $B$ is rescaled to sum up to $B$ such that the average weight for each example is $1$. With this weighting mechanism, the preliminary definition of \emph{class confusion} is
\begin{equation}\label{eq:CHC}
   {{\mathbf{C}}_{jj'}} = {\widehat{\mathbf{y}}}_{ \cdot j}^{\sf T}{\mathbf{W}}{{\widehat{\mathbf{y}}}_{ \cdot j'}}.
\end{equation}

\textbf{Category Normalization.}
The batch-based definition of the class confusion in Eq.\eqref{eq:CHC} is native for the mini-batch SGD optimization. However, when the number of classes is large, it will run into a severe \textit{class imbalance} in each batch.
To tackle this problem, we adopt a category normalization technique widely used in Random Walk \cite{vonLuxburg2007}:
\begin{equation}\label{eq:norm}
   {{{\widetilde{\mathbf C}}}_{jj'}} = \frac{{{{\mathbf{C}}_{jj'}}}}{{\sum\nolimits_{{j''} = 1}^{|{\mathcal{C}}|} {{{\mathbf{C}}_{j{j''}}}} }}.
\end{equation}
Taking the idea of Random Walk, the normalized class confusion in Eq.\eqref{eq:norm} has a neat interpretation: It is probable to walk from one class to another (resulting in the wrong classification) if the two classes have a high class confusion.

\textbf{Minimum Class Confusion.}
Given the aforementioned derivations, we can formally define the loss function to enable Versatile Domain Adaptation (VDA). Recall that ${{{\widetilde{\mathbf C}}}_{jj'}}$ well measures the confusion between each class pair $j$ and $j'$. We only need to minimize the cross-class confusion, \emph{i.e.} $j\ne j'$. Namely, the ideal situation is that no examples are ambiguously classified into two classes at the same time. In this sense, the {Minimum Class Confusion (MCC)} loss is defined as
\begin{equation}\label{eq:MCC}
   {L_{{\rm{MCC}}}} ( {{{\widehat {\mathbf{Y}}}_t}} ) = \frac{1}{|{\cal {C}}|}\sum\limits_{j = 1}^{|{\mathcal{C}}|} {\sum\limits_{j' \ne j}^{|{\mathcal{C}}|} {\left| {{{{\widetilde{\mathbf C}}}_{jj'}}} \right|} }.
\end{equation}
Since the class confusion in Eq.~\eqref{eq:norm} has been normalized, minimizing the \emph{between-class} confusion in Eq.~\eqref{eq:MCC} implies that the \emph{within-class} confusion is maximized. Note that Eq.~\eqref{eq:MCC} is a general loss that is pluggable to existing approaches.

We want to emphasize that the inductive bias of \emph{class confusion} in this work is more general than that of \emph{domain alignment} in previous work \cite{DANN2016,Long15DAN,JAN2017,Long18CDAN,MCD2018}. As discussed in Section \ref{relate}, many previous methods explicitly align features from the source and target domains, at the risk of deteriorating the feature discriminability and impeding the transferability~\cite{BSP2019}. Further, the inductive bias of \emph{class confusion} is general and applicable to a variety of domain adaptation scenarios, while that of \emph{domain alignment} will suffer when the domains cannot be aligned naturally (\emph{e.g.} the partial-set DA scenarios \cite{SAN2018,Cao2017PADA,IWAN2018}).

\subsection{Versatile Approach to Domain Adaptation}

The main motivation of this work is to design a versatile approach to a variety of DA scenarios. As the class confusion is a common inductive bias of many DA scenarios, combining the cross-entropy loss on the source labeled data and the MCC loss on the target unlabeled data will enable these DA scenarios. 

Denote by ${\widehat {\mathbf{y}}_s} = G(F({{\mathbf{x}}_s}))$ the class prediction for a source example ${{\mathbf{x}}_s}$, and by ${\widehat {\mathbf{Y}}_t} = G(F({{\mathbf{X}}_t}))$ the class predictions for a batch of $B$ target examples ${{\mathbf{X}}_t}$. The versatile approach (also termed by \textbf{MCC} for clarity) proposed in this paper for a variety of domain adaptation scenarios is formulated as
\begin{equation}\label{Method}
   \mathop {\min }\limits_{F,G} \, {\mathbb{E}_{({{\mathbf{x}}_s},{{\mathbf{y}}_s}) \in \mathcal{S}}}{L_{{\rm{CE}}}}\left( {{{\widehat {\mathbf{y}}}_s},{{\mathbf{y}}_s}} \right) + \mu \, {\mathbb{E}_{{{\mathbf{X}}_t} \subset \mathcal{T}}}{L_{{\rm{MCC}}}}( {{{\widehat {\mathbf{Y}}}_t}} ),
\end{equation} 
where $L_{\rm CE}$ is the cross-entropy loss and $\mu$ is a hyper-parameter for the MCC loss. With this joint loss, feature extractor $F$ and classifier $G$ of the deep DA model can be trained end-to-end by back-propagation. Note that, Eq.~\eqref{Method} is a \emph{versatile} approach to many DA scenarios without any modifications to the loss.
\begin{itemize}
\item \textbf{Unsupervised Domain Adaptation (UDA).} Since Eq.~\eqref{Method} is formulated natively for this vanilla domain adaptation scenario,  MCC can be directly applied to this scenario without any modification.

\item \textbf{Partial Domain Adaptation (PDA).} Without explicit domain alignment,  we need not to worry about the \emph{misalignment} between source outlier classes and target classes, which is the technical bottleneck of PDA \cite{Cao2017PADA}. Meanwhile, compared to the confusion between the target classes, the confusion between the source outlier classes on the target domain is negligible in the MCC loss.
Therefore, we can directly apply Eq.~\eqref{Method} to PDA. 

\item \textbf{Multi-Source Domain Adaptation (MSDA).} Prior methods of MSDA consider multiple source domains as different domains, capturing the internal source domain shifts, and a simple merge of source domains proves fragile. However, since MCC is based on class confusion instead of domain alignment, we can safely merge $S$ source domains as $\mathcal{S} \leftarrow {\mathcal{S}_1} \cup  \cdots  \cup {\mathcal{S}_S}$.

\item \textbf{Multi-Target Domain Adaptation (MTDA).} Though a simple merge of target domains is risky for existing methods, for MCC applied to MTDA, we can safely merge $T$ target domains as $\mathcal{T} \leftarrow {\mathcal{T}_1} \cup  \cdots  \cup {\mathcal{T}_T}$.

\item \textbf{Multi-Source/Multi-Target Partial Domain Adaptation (MSPDA / MTPDA).} As MCC can directly tackle PDA and MSDA/MTDA, it can handle these derived scenarios by simply merging multiple sources or targets.
\end{itemize}

\subsection{Regularizer to Existing DA Methods}

Since the inductive bias of \emph{class confusion} is orthogonal to the widely-used \emph{domain alignment}, our method is well complementary to the previous methods. The MCC loss Eq.~\eqref{eq:MCC} can serve as a regularizer pluggable to existing methods.

We take as an example the standard domain alignment framework \cite{DANN2016,Long18CDAN} based on domain-adversarial training. Integrating the MCC loss as a regularizer yields
\begin{equation}
   \centering
   \label{regularization}
   \begin{aligned}
      \mathop {\min }\limits_{F,G} \mathop {\max }\limits_D \; {\mathbb{E}_{({{\mathbf{x}}_s},{{\mathbf{y}}_s}) \in \mathcal{S}}}{L_{{\rm{CE}}}}\left( {{{\widehat {\mathbf{y}}}_s},{{\mathbf{y}}_s}} \right) + \mu \, {\mathbb{E}_{{{\mathbf{X}}_t} \subset \mathcal{T}}}{L_{{\rm{MCC}}}}({\widehat {\mathbf{Y}}_t}) - \lambda \, {\mathbb{E}_{{\mathbf{x}} \in \mathcal{S} \cup \mathcal{T}}}{L_{{\rm{CE}}}}(D( {\widehat {\mathbf{f}}} ),{\bf d}) ,
   \end{aligned}
\end{equation}
where the third term is the domain-adversarial loss for the domain discriminator $D$ striving to distinguish the source from the target, and ${\bf d}$ is the domain label, $\widehat{\mathbf{f}} = F(\mathbf{x})$ is the feature representation learned to confuse the domain discriminator. The overall framework is a \emph{minimax} game between two players $F$ and $D$, in which $\lambda$ and $\mu$ are trade-off hyper-parameters between different loss functions.
Generally, the MCC loss is also readily pluggable to other representative domain adaptation frameworks, \emph{e.g.} moment matching \cite{Long15DAN} and large norm \cite{AFN2019}.

\section{Experiments}\label{sec:exp}

We evaluate MCC as a standalone approach with many methods for six domain adaptation scenarios (UDA, MSDA, MTDA, PDA, MSPDA and MTPDA). We also evaluate MCC as a regularizer to existing domain adaptation methods.

\subsection{Setup}

We use four standard datasets: \textbf{(1) Office-31}~\cite{Saenko10Office}: a classic domain adaptation dataset with 31 categories and 3 domains: \textit{Amazon}~(\textbf{A}), \textit{Webcam}~(\textbf{W}) and \textit{DSLR}~(\textbf{D}); \textbf{(2) Office-Home}~\cite{Venkateswara17Officehome}: a more difficult dataset (larger domain shift) with 65 categories and 4 domains: \textit{Art}~(\textbf{A}), \textit{Clip Art}~(\textbf{C}), \textit{Product}~(\textbf{P}) and \textit{Real World}~(\textbf{R}); \textbf{(3) VisDA-2017}~\cite{peng2017VisDA}: a dataset with 12 categories and over 280,000 images; \textbf{(4) DomainNet}~\cite{peng2018moment}: the largest and hardest domain adaptation dataset, with approximately 0.6 million images from 345 categories and 6 domains: \textit{Clipart} (\textbf{c}), \textit{Infograph} (\textbf{i}), \textit{Painting} (\textbf{p}), \textit{Quickdraw} (\textbf{q}), \textit{Real} (\textbf{r}) and \textit{Sketch} (\textbf{s}). 

Our methods are implemented based on \textbf{PyTorch}. ResNet~\cite{he2016Resnet} pre-trained on ImageNet~\cite{deng2009imagenet} is used as the network backbone. 
We use Deep Embedded Validation (DEV)~\cite{you19icml} to select hyper-parameter $T$ and provide parameter sensitivity analysis. A balance between the cross-entropy and MCC, \emph{i.e.} $\mu=1.0$ works well for all experiments. We run each experiment for $5$ times and report the average results.

\subsection{Results and Discussion}

\textbf{Multi-Target Domain Adaptation (MTDA).}
We evaluate the MTDA tasks following the protocol of DADA~\cite{Peng2019DADA}, which provides six tasks on DomainNet, the most difficult dataset to date. We adopt the strategy that directly merges multiple target domains. As shown in Table~\ref{table:domainnet}, many competitive methods are not effective in this challenging scenario. However, our simple method outperforms the current state-of-the-art method DADA~\cite{Peng2019DADA} by a big margin ($\mathbf{7.3\%}$). Note that the source-only accuracy is rather low on this dataset, validating that our method, with well-designed mechanisms, is sufficiently robust to wrong predictions.

\textbf{Multi-Source Domain Adaptation (MSDA).}
When running our method for MSDA, we similarly merge multiple source domains in MCC and compare it to existing DA algorithms that are specifically designed for MSDA on DomainNet. As shown in Table \ref{table:domainnet}, based on the inductive bias of minimizing the class confusion, MCC significantly outperforms $\rm M^{3}SDA$~\cite{peng2018moment}, the state-of-the-art method  by a big margin ($\mathbf{5.0\%}$). Note that these specific methods are of very complex architecture and loss designs and may be hard to use in practical applications.

\begin{table}[htbp]
   \addtolength{\tabcolsep}{-0.5pt}
   \scriptsize
   \caption{Accuracy (\%) on {DomainNet} for MTDA and MSDA (ResNet-101).}
   \label{table:domainnet}
   \subtable[MTDA]{
   \begin{tabular}{lcccccccc}
      \toprule
      Method      & c: & i: & p: & q: & r: & s: & Avg  \\
      \midrule
      ResNet~\cite{he2016Resnet} & 25.6  & 16.8   & 25.8 & 9.2  & 20.6  & 22.3   & 20.1 \\
      SE~\cite{SE2018} & 21.3 & 8.5 & 14.5 & 13.8 & 16.0 & 19.7 & 15.6 \\
      MCD~\cite{MCD2018}& 25.1 & 19.1 & 27.0 & 10.4 & 20.2 & 22.5 & 20.7 \\
      DADA~\cite{Peng2019DADA} &26.1 & 20.0 & 26.5 & 12.9 & 20.7 & 22.8 & 21.5\\
      \midrule
      \textbf{MCC}   &\textbf{33.6} & \textbf{30.0} & \textbf{32.4} &\textbf{13.5} & \textbf{28.0} & \textbf{35.3} & \textbf{28.8}     \\
      \bottomrule
   \end{tabular}
   }
   \subtable[MSDA]{
   \begin{tabular}{lccccccc}
      \toprule
      Method       &:c & :i & :p & :q & :r & :s & Avg  \\
      \midrule
      ResNet~\cite{he2016Resnet} & 47.6          & 13.0          & 38.1          & 13.3          & 51.9          & 33.7          & 32.9 \\
      % DAN        & 45.4          & 12.8          & 36.2          & 15.3          & 48.6          & 34.0          & 32.1 \\
      % RTN         & 44.2          & 12.6          & 35.3          & 14.6          & 48.4          & 31.7          & 31.1 \\
      % JAN  & 40.9 & 11.1 & 35.4 & 12.1 & 45.8 & 32.3 & 29.6\\
      
      % ADDA & 47.5 & 11.4 & 36.7 & 14.7 & 49.1 & 33.5 & 32.2 \\
      % SE & 24.7 & 3.9 & 12.7 & 7.1 & 22.8 & 9.1 & 16.1 \\
      MCD~\cite{MCD2018} & 54.3 & 22.1 & 45.7 & 7.6 & 58.4 & 43.5 & 38.5 \\      
      DCTN~\cite{xu2018deep}        & 48.6          & 23.5          & 48.8          & 7.2           & 53.5          & 47.3          & 38.2 \\
      M$^{3}$SDA~\cite{peng2018moment}       & 58.6          & 26.0          & 52.3          & 6.3           & 62.7          & 49.5          & 42.6 \\
      \midrule
      \textbf{MCC}      & \textbf{65.5}          & \textbf{26.0}          &  \textbf{56.6}          &  \textbf{16.5}             &  \textbf{68.0}            &  \textbf{52.7}            &  \textbf{47.6}   \\
      \bottomrule
   \end{tabular}
   }
\end{table}

\textbf{Partial Domain Adaptation (PDA).} Due to the existence of source outlier classes, PDA is known as a challenging scenario because of the misalignment between the source and target classes. For a fair comparison, we follow the protocol of PADA~\cite{Cao2017PADA} and AFN~\cite{AFN2019}, where the first $25$ categories (in alphabetic order) of the Office-Home dataset are taken as the target domain. As shown in Table~\ref{table:office-home-pda}, on this dataset, MCC outperforms AFN~\cite{AFN2019}, the \emph{ICCV'19 honorable-mention} entry and the state-of-the-art method for PDA, by a big margin ($\mathbf{3.3\%}$).

\begin{table}[htbp]
   \addtolength{\tabcolsep}{1pt}
   \centering
   \scriptsize
   \caption{Accuracy (\%) on {Office-Home} for PDA  (ResNet-50).}
   \label{table:office-home-pda}
   \begin{tabulary}{\linewidth}{l*{14}Cc}
      \toprule
      Method (S:T) & A:C & A:P & A:R & C:A & C:P & C:R & P:A & P:C & P:R & R:A & R:C & R:P & Avg \\
      \midrule
      ResNet~\cite{he2016Resnet} & 38.6 & 60.8 & 75.2 & 39.9 & 48.1 & 52.9 & 49.7 & 30.9 & 70.8 & 65.4 & 41.8 & 70.4 & 53.7 \\
      DAN~\cite{Long15DAN} & 44.4 & 61.8 & 74.5 & 41.8 & 45.2 & 54.1 & 46.9 & 38.1 & 68.4 & 64.4 & 51.5 & 74.3 & 56.3 \\
      JAN~\cite{Long17JAN} & 45.9 & 61.2 & 68.9 & 50.4 & 59.7 & 61.0 & 45.8 & 43.4 & 70.3 & 63.9 & 52.4 & 76.8 & 58.3 \\
      PADA~\cite{Cao2017PADA} & 51.2 & 67.0 & 78.7 & 52.2 & 53.8 & 59.0 & 52.6 & 43.2 & 78.8 & 73.7 & 56.6 & 77.1 & 62.0   \\
      AFN~\cite{AFN2019} & 58.9 & 76.3 & 81.4 & 70.4 & \textbf{73.0} & 77.8 & 72.4 & 55.3 & 80.4 & 75.8 & 60.4 & 79.9 & 71.8\\
      \midrule
      \textbf{MCC}  &  \textbf{63.1} & \textbf{80.8} & \textbf{86.0} & \textbf{70.8} & 72.1 & \textbf{80.1} & \textbf{75.0} & \textbf{60.8} &  \textbf{85.9} & \textbf{78.6} & \textbf{65.2} & \textbf{82.8} & \textbf{75.1}\\
      \bottomrule
   \end{tabulary}
\end{table}

\textbf{Unsupervised Domain Adaptation (UDA).} We evaluate MCC for the most common UDA scenario on several datasets. {(1)} \textit{VisDA-2017}. As reported in Table \ref{table:VisDA}, 
MCC surpasses state-of-the-art UDA algorithms and yields the highest accuracy to date among methods of no complex architecture and loss designs. 
{(2)} \textit{Office-31}. As shown in Table \ref{table:office31}, MCC performs the best.
(3) \textit{Two Moon} \cite{TAT2019}. We train a shallow MLP from scratch and plot the decision boundaries of MCC and Minimum Entropy (MinEnt) \cite{Grandvalet2005Semi}. MCC yields much better boundaries in Fig.~\ref{twomoon}.

\begin{table}[h]
   \addtolength{\tabcolsep}{1pt}
   \centering
   \scriptsize
   \caption{Accuracy (\%) on {VisDA-2017} for UDA (ResNet-101).}
   \label{table:VisDA}
   \begin{tabulary}{\linewidth}{l*{14}Cc}
      \toprule
      Method  & plane & bcybl & bus & car & horse & knife & mcyle & persn & plant & sktb & train & truck & mean \\
      \midrule
      ResNet \cite{he2016Resnet} & 55.1& 53.3 & 61.9 & 59.1 & 80.6 & 17.9 & 79.7 & 31.2 & 81.0 & 26.5 & 73.5 & 8.5 & 52.4 \\
      MinEnt \cite{Grandvalet2005Semi} & 80.3 & 75.5 & 75.8 & 48.3 & 77.9 & 27.3 & 69.7 & 40.2 & 46.5 & 46.6 & 79.3 & 16.0 & 57.0 \\
      DANN~\cite{DANN2016} & 81.9 & 77.7 & 82.8 & 44.3 & 81.2 & 29.5 &  65.1 & 28.6 & 51.9 & 54.6 & 82.8 & 7.8 & 57.4\\
      DAN \cite{Long15DAN} & 87.1 & 63.0 & 76.5 & 42.0 & 90.3 & 42.9 & 85.9 & 53.1 & 49.7 & 36.3 & 85.8 & 20.7 & 61.1 \\
      MCD~\cite{MCD2018} & 87.0 & 60.9 & 83.7 & 64.0 & 88.9 & 79.6 & 84.7 & 76.9 & 88.6 & 40.3 & 83.0 & 25.8 & 71.9 \\
      CDAN~\cite{Long18CDAN}&  85.2 & 66.9 &  83.0 & 50.8 & 84.2 & 74.9 & 88.1 & 74.5 & 83.4 & \textbf{76.0} & 81.9 & \textbf{38.0} & 73.9\\
      % ADR~\cite{ADR2018} & 87.8 & 79.5 & 83.7 & 65.3 & 92.3 & 61.8 & 88.9 & 73.2 & 87.8 & 60.0 & 85.5 & 32.3 & 74.8 \\
      AFN~\cite{AFN2019} & \textbf{93.6} & 61.3 & \textbf{84.1} & 70.6 & \textbf{94.1} & 79.0 & \textbf{91.8} & \textbf{79.6} & \textbf{89.9} & 55.6 & \textbf{89.0} & 24.4 & 76.1 \\
      \midrule
      \textbf{MCC} & 88.1 & \textbf{80.3} & 80.5 & \textbf{71.5} & 90.1 & \textbf{93.2} & 85.0 & 71.6 & 89.4 & 73.8 & 85.0 & 36.9 &\textbf{78.8} \\
      \bottomrule
   \end{tabulary}
\end{table}

\begin{table}[h]
   \addtolength{\tabcolsep}{1pt}
   \centering
   \caption{Accuracy (\%) on {Office-31} for UDA (ResNet-50).}
   \scriptsize
   \label{table:office31}
   \begin{tabulary}{\linewidth}{l*{7}c}
      \toprule
      Method & A$\rightarrow$W & D$\rightarrow$W & W$\rightarrow$D & A$\rightarrow$D & D$\rightarrow$A & W$\rightarrow$A & Avg \\
      \midrule
      ResNet \cite{he2016Resnet} & 68.4$\pm$0.2 & 96.7$\pm$0.1 & 99.3$\pm$0.1 & 68.9$\pm$0.2 & 62.5$\pm$0.3 & 60.7$\pm$0.3 & 76.1 \\
      DAN \cite{Long15DAN} \ & 80.5$\pm$0.4 & 97.1$\pm$0.2 & 99.6$\pm$0.1 & 78.6$\pm$0.2 & 63.6$\pm$0.3 & 62.8$\pm$0.2 & 80.4 \\
      RTN \cite{Long16RTN} & 84.5$\pm$0.2 & 96.8$\pm$0.1 & 99.4$\pm$0.1 & 77.5$\pm$0.3 & 66.2$\pm$0.2 & 64.8$\pm$0.3 & 81.6 \\
      DANN \cite{DANN2016}  & 82.0$\pm$0.4 & 96.9$\pm$0.2 & 99.1$\pm$0.1 & 79.7$\pm$0.4 & 68.2$\pm$0.4 & 67.4$\pm$0.5 & 82.2 \\
      % ADDA \cite{ADDA2017} & 86.2 $\pm$ 0.5 & 96.2 $\pm$ 0.3 & 98.4 $\pm$ 0.3 & 77.8 $\pm$ 0.3 & 69.5 $\pm$ 0.4 & 68.9 $\pm$ 0.5 & 82.9 \\
      JAN \cite{Long17JAN} & 85.4$\pm$0.3 & 97.4$\pm$0.2 & 99.8$\pm$0.2 & 84.7$\pm$0.3 & 68.6$\pm$0.3 & 70.0$\pm$0.4 & 84.3 \\
      % MADA \cite{Pei18MADA} & 90.0$\pm$0.1 & 97.4$\pm$0.1 & 99.6$\pm$0.1 & 87.8$\pm$0.2 & 70.3$\pm$0.3 & 66.4$\pm$0.3 & 85.2 \\
      % MinEnt \cite{Grandvalet2005Semi} & 92.5$\pm$0.4 & 98.0$\pm$0.2 & 99.8$\pm$0.2 & 92.6$\pm$0.3 & 70.3$\pm$0.2 & 63.1$\pm$0.2 & 86.1 \\
      % SimNet \cite{Pedro17SimNet} & 88.6$\pm$0.5 & 98.2$\pm$0.2 & 99.7$\pm$0.2 & 85.3$\pm$0.3 & 73.4$\pm$0.8 & 71.6$\pm$0.6 & 86.2 \\
      GTA \cite{Swami17GTA} & 89.5$\pm$0.5 & 97.9$\pm$0.3 & 99.8$\pm$0.4 & 87.7$\pm$0.5 & 72.8$\pm$0.3 & 71.4$\pm$0.4 & 86.5 \\
      CDAN \cite{Long18CDAN} & 94.1$\pm$0.1 & \textbf{98.6}$\pm$0.1 & \textbf{100.0}$\pm$0.0 & 92.9$\pm$0.2 & 71.0$\pm$0.3 & 69.3$\pm$0.3 & 87.7 \\
      AFN \cite{AFN2019}& 88.8$\pm$0.5 & 98.4$\pm$0.3 & 99.8$\pm$0.1 & 87.7$\pm$0.6 & 69.8$\pm$0.4 & 69.7$\pm$0.4 & 85.7 \\
      MDD \cite{MDD2019}& 94.5$\pm$0.3 & 98.4$\pm$0.3 & \textbf{100.0}$\pm$0.0 & 93.5$\pm$0.2 & \textbf{74.6}$\pm$0.3 & 72.2$\pm$0.1 & 88.9 \\
      \midrule
      \textbf{MCC} & \textbf{95.5}$\pm$0.2 & \textbf{98.6}$\pm$0.1 & \textbf{100.0}$\pm$0.0 & \textbf{94.4}$\pm$0.3 & 72.9$\pm$0.2 & \textbf{74.9}$\pm$0.3 & \textbf{89.4} \\
      \bottomrule
   \end{tabulary}
\end{table}

\begin{figure}[!htbp]
   \centering
   \subfigure[\tiny MinEnt~\cite{Grandvalet2005Semi}]{
      \includegraphics[width=0.20\textwidth]{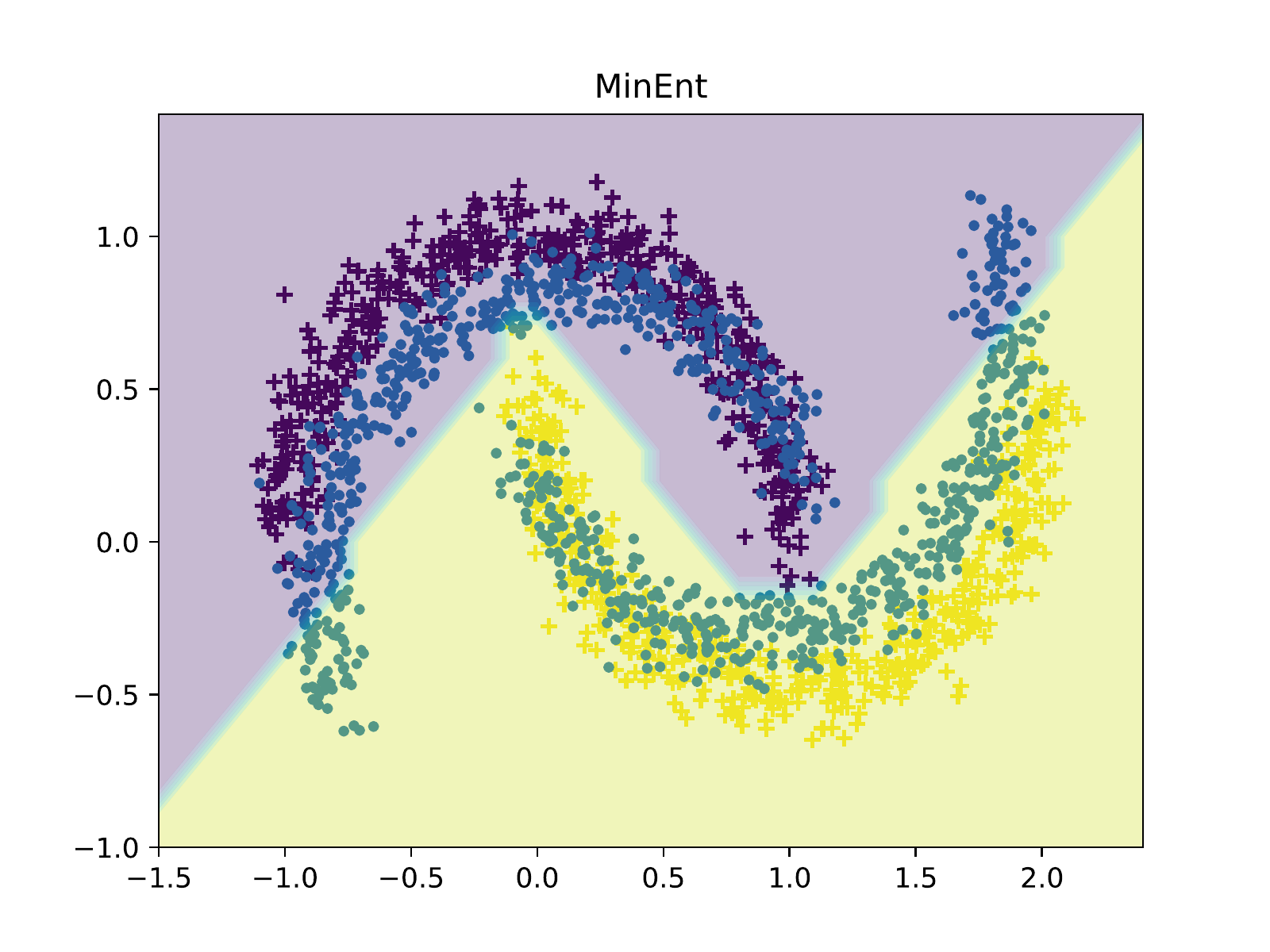}
      \label{twomoon_MinEnt}
   }
   \hfil
   \subfigure[\tiny MCC]{
      \includegraphics[width=0.20\textwidth]{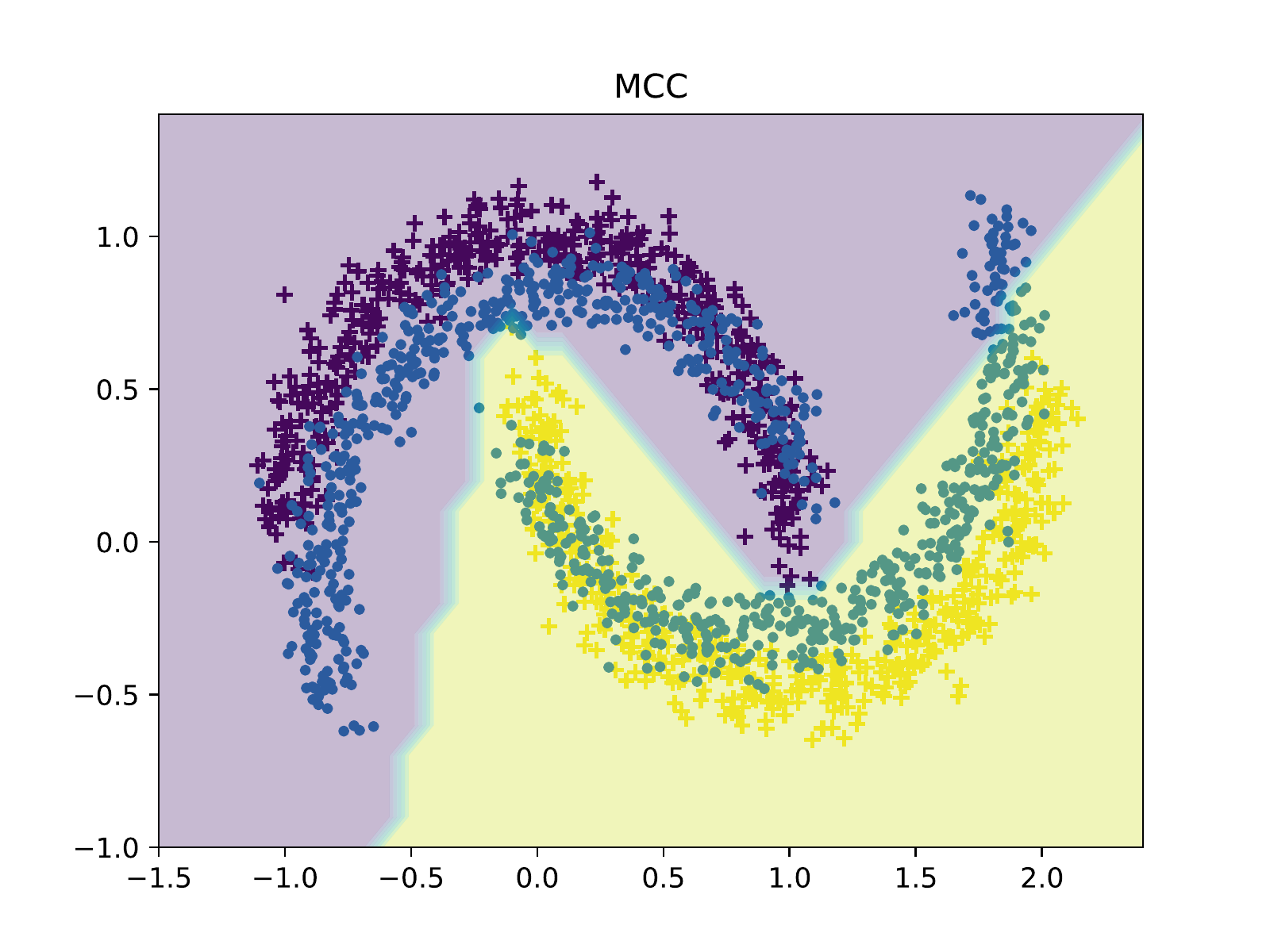}
      \label{twomoon_MCC}
   }
   \hfil
   \subfigure[\tiny DANN+MinEnt]{
      \includegraphics[width=0.20\textwidth]{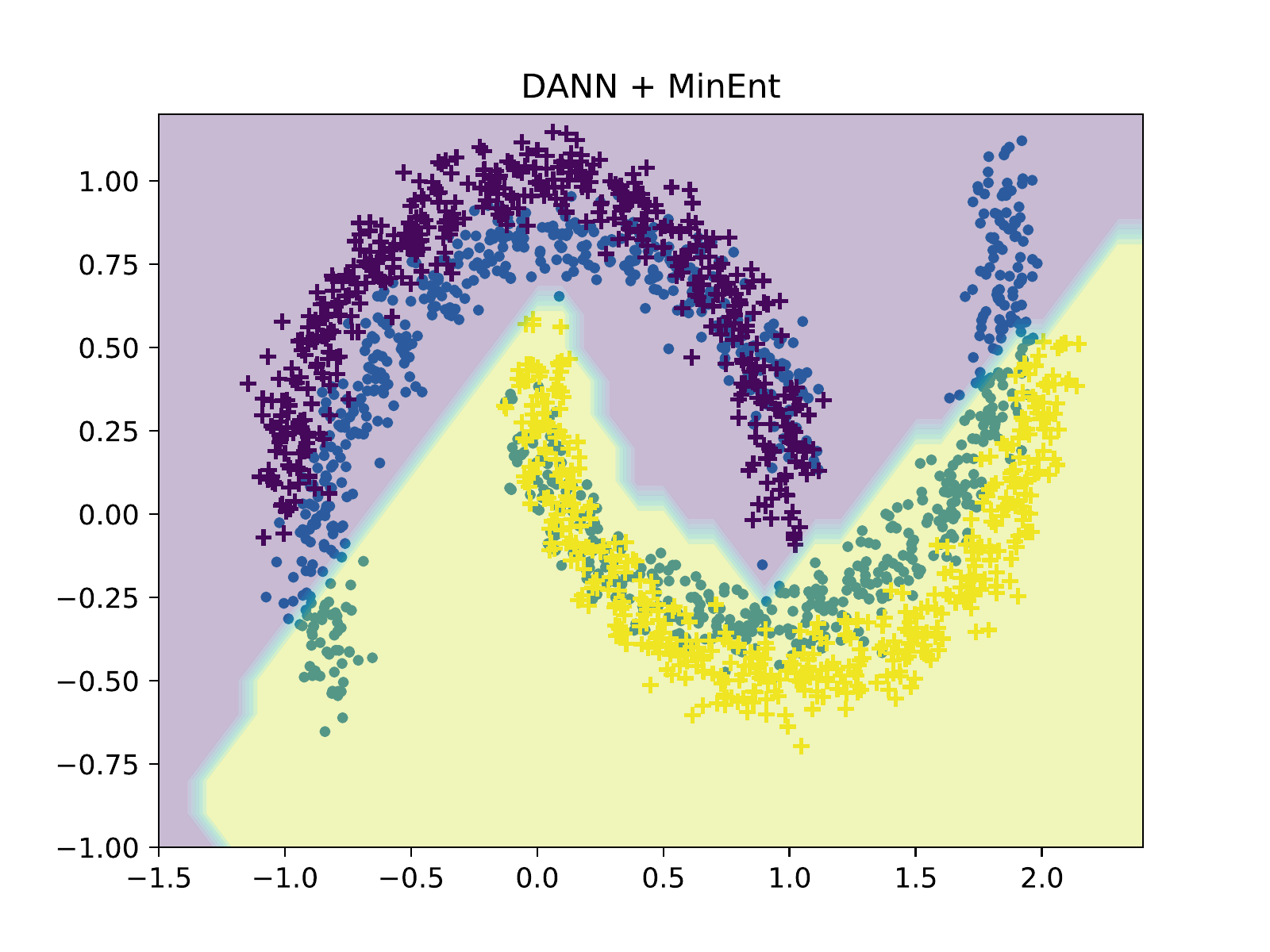}
      \label{twomoon_DANN_MinEnt}
   }
   \hfil
   \subfigure[\tiny DANN+MCC]{
      \includegraphics[width=0.20\textwidth]{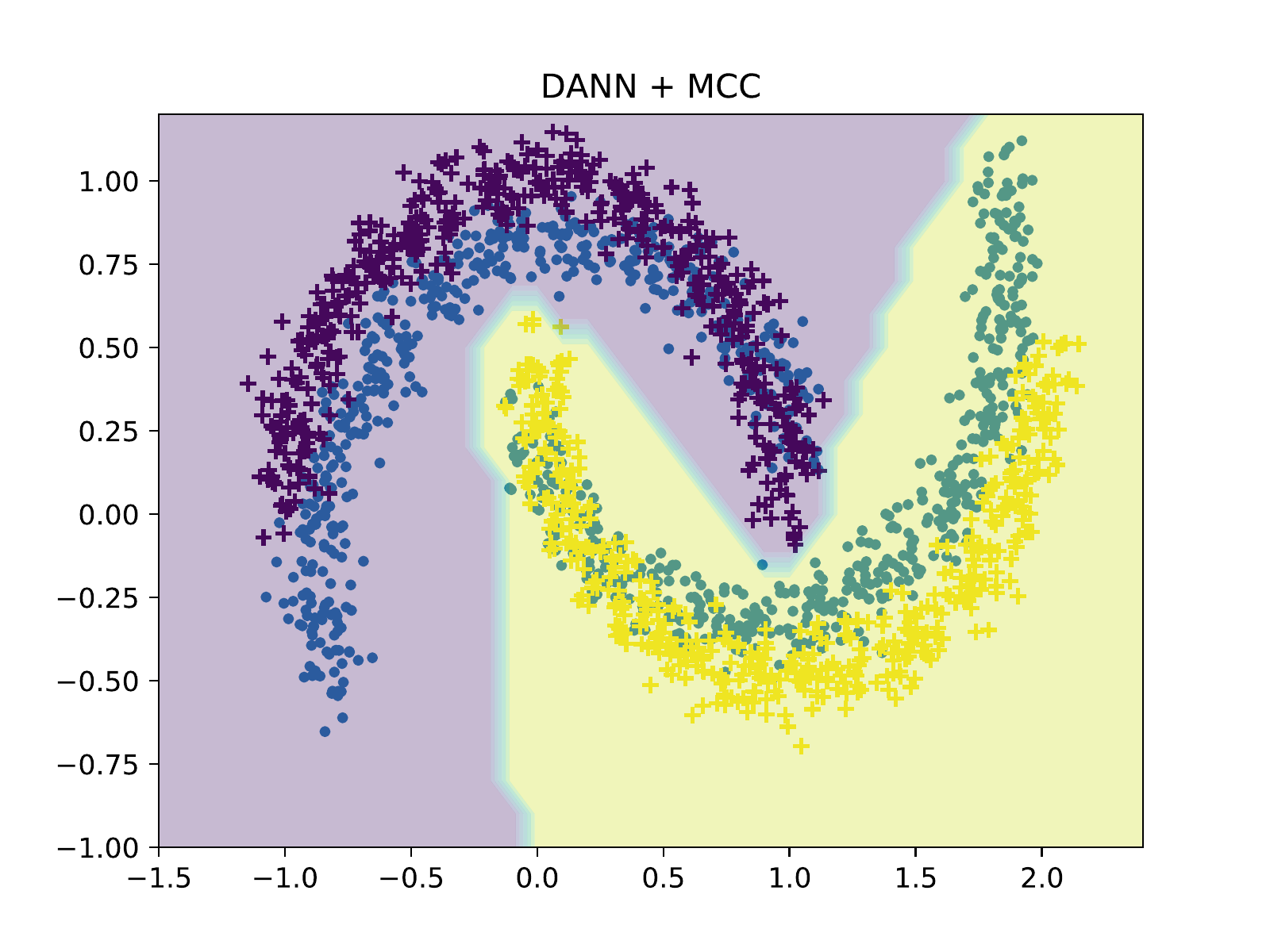}
      \label{twomoon_DANN_MCC}
   }
   \caption{Decision boundaries on the Two Moon dataset. Blue points indicate target data, and different classes of the source data are depicted in purple and yellow.}
   \label{twomoon}
\end{figure}

\textbf{Multi-Source/Multi-Target Partial Domain Adaptation (MSPDA / MTPDA).} Table~\ref{tab:MSPDA} shows that MCC is versatile to handle these hard scenarios.

\begin{table}[!htbp]
   \addtolength{\tabcolsep}{2pt}
   \scriptsize
   \caption{Accuracy (\%) on Office-Home for MSPDA and MTPDA.}
   \label{tab:MSPDA}
   \centering
   \subtable[MSPDA]{
   \begin{tabular}{lccccc}
      \toprule
         Method & :A & :C & :P & :R  & Avg \\
      \midrule
         DANN~\cite{DANN2016}   & 58.3& 43.6 & 60.7 & 71.2  & 58.5 \\
         PADA~\cite{Cao2017PADA}   & 62.8& 51.8 & 71.7 & 79.2  & 66.4 \\
         $\rm M^{3}SDA$~\cite{peng2018moment} & 67.4 & 55.3 & 72.2 & 80.4 & 68.8\\
         AFN~\cite{AFN2019}    & 77.1& 61.2 & 79.3 & 82.5  & 75.0 \\
      \midrule
      \textbf{MCC}    & \textbf{79.6}& \textbf{67.5} & \textbf{80.6} & \textbf{85.1}  & \textbf{78.2} \\
      \bottomrule
   \end{tabular}
   }
   \hfil
   \subtable[MTPDA]{
   \begin{tabular}{lccccccc}
      \toprule
         Method & A: & C: & P: & R: & Avg \\
      \midrule
         DANN~\cite{DANN2016}   & 44.6 & 44.8& 39.1 & 44.1  & 43.1 \\
         PADA~\cite{Cao2017PADA}   & 59.9 & 53.7& 51.1 & 61.4  & 56.5 \\
         DADA~\cite{Peng2019DADA}   & 65.1 & 63.0& 60.4 & 63.0  & 62.9 \\
         AFN~\cite{AFN2019}    & 68.7 & 65.6& 63.4 & 67.5  & 66.3 \\
         \midrule
         \textbf{MCC}    & \textbf{73.1} & \textbf{72.1}& \textbf{69.4} & \textbf{68.3}  & \textbf{70.7} \\
      \bottomrule
   \end{tabular}
   }
\end{table}

\subsection{Empirical Analyses}

\textbf{General Regularizer.} MCC can be used as a general regularizer for existing DA methods. We compare its performance with entropy minimization (MinEnt) \cite{Grandvalet2005Semi} and Batch Spectral Penalization (BSP) \cite{BSP2019}. As shown in Tables~\ref{table:VisDA-reg} and \ref{table:office31-reg}, MCC yields larger improvements than MinEnt and BSP to a variety of DA methods.

\begin{table}[!htbp]
   \centering
   \caption{Accuracy (\%) on {VisDA-2017} as \textit{regularizer} for UDA (ResNet-101).}
   \label{table:VisDA-reg}
   \scriptsize
   \begin{tabulary}{\linewidth}{l*{14}Cc}
      \toprule
      Method  & plane & bcybl & bus & car & horse & knife & mcyle & persn & plant & sktb & train & truck & mean \\
      \midrule
      DANN~\cite{DANN2016} & 81.9 & 77.7 & 82.8 & 44.3 & 81.2 & 29.5 &  65.1 & 28.6 & 51.9 & 54.6 & 82.8 & 7.8 & 57.4\\
      DANN + MinEnt~\cite{Grandvalet2005Semi}    & 87.4 & 55.0 & 75.3 & \textbf{63.8} & 87.4 & 43.6 & \textbf{89.3} & 72.5 & 82.9 & \textbf{78.6} & 85.6 & 27.4 & 70.7 \\
      DANN + BSP~\cite{BSP2019} & \textbf{92.2} & 72.5 & \textbf{83.8} & 47.5 & 87.0 & 54.0 & 86.8 & 72.4 & 80.6 & 66.9 & 84.5 & 37.1 & 72.1 \\
      \textbf{DANN + MCC} & 90.4 & \textbf{79.8} & 72.3 & 55.1 & \textbf{90.5} &  \textbf{86.8} & 86.6 & \textbf{80.0} & \textbf{94.2} & 76.9 & \textbf{90.0} & \textbf{49.6} & \textbf{79.4}\\
      \midrule
      CDAN~\cite{Long18CDAN}&  85.2 & 66.9 &  \textbf{83.0} & 50.8 & 84.2 & 74.9 & 88.1 & 74.5 & 83.4 & 76.0 & 81.9 & 38 & 73.9\\
      CDAN + MinEnt~\cite{Grandvalet2005Semi}    & 90.5 & 65.8 & 79.1 & 62.2 & 89.8 & 28.7 &  \textbf{92.8} & 75.4& 86.8 & 65.3 & 85.2 & 35.3 & 71.4 \\
      CDAN + BSP~\cite{BSP2019} & 92.4 & 61.0 & 81.0 & 57.5 & 89.0 &  \textbf{80.6} & 90.1 & 77.0 & 84.2 & 77.9 & 82.1 & 38.4 & 75.9 \\
      \textbf{CDAN + MCC} &   \textbf{94.5} &  \textbf{80.8} & 78.4 &  \textbf{65.3} &  \textbf{90.6} & 79.4 & 87.5 &  \textbf{82.2} &  \textbf{94.7} &  \textbf{81.0}&  \textbf{86.0} &  \textbf{44.6} &   \textbf{80.4}\\
      \bottomrule
   \end{tabulary}
\end{table}

\begin{table}[!htbp]
   \addtolength{\tabcolsep}{1pt}
   \centering
   \caption{Accuracy (\%) on {Office-31} as \textit{regularizer} for UDA (ResNet-50).}
   \label{table:office31-reg}
   \scriptsize
   \begin{tabulary}{\linewidth}{l*{7}c}
      \toprule
      Method & A$\rightarrow$W & D$\rightarrow$W & W$\rightarrow$D & A$\rightarrow$D & D$\rightarrow$A & W$\rightarrow$A & Avg \\
      \midrule
      DANN~\cite{DANN2016} & 82.0$\pm$0.4 & 96.9$\pm$0.2 & 99.1$\pm$0.1 & 79.7$\pm$0.4 & 68.2$\pm$0.4 & 67.4$\pm$0.5 & 82.2 \\
      DANN + MinEnt~\cite{Grandvalet2005Semi} & 91.7$\pm$0.3 & 98.3$\pm$0.1 & \textbf{100.0}$\pm$0.0 & 87.9$\pm$0.3 & 68.8$\pm$0.3 & 68.1$\pm$0.3 & 85.8   \\
      DANN + BSP~\cite{BSP2019} & 93.0$\pm$0.2 & 98.0$\pm$0.2 & \textbf{100.0}$\pm$0.0 & 90.0$\pm$0.4 & 71.9$\pm$0.3 & 73.0$\pm$0.3 & 87.7   \\
      \textbf{DANN + MCC}  & \textbf{95.6}$\pm$0.3 & \textbf{98.6}$\pm$0.1 & 99.3$\pm$0.0 & \textbf{93.8}$\pm$0.4 & \textbf{74.0}$\pm$0.3 & \textbf{75.0}$\pm$0.4 & \textbf{89.4} \\
      \midrule
      CDAN~\cite{Long18CDAN} & 94.1$\pm$0.1 & \textbf{98.6}$\pm$0.1 & \textbf{100.0}$\pm$0.0 & 92.9$\pm$0.2 & 71.0$\pm$0.3 & 69.3$\pm$0.3 & 87.7 \\
      CDAN + MinEnt~\cite{Grandvalet2005Semi}  & 91.7$\pm$0.2 & 98.5$\pm$0.1 & \textbf{100.0}$\pm$0.0 & 90.4$\pm$0.3 & 72.3$\pm$0.2 & 69.5$\pm$0.2 & 87.1 \\ 
      CDAN + BSP~\cite{BSP2019} &   93.3$\pm$0.2 & 98.2$\pm$0.2 & \textbf{100.0}$\pm$0.0 & 93.0$\pm$0.2 & \textbf{73.6}$\pm$0.3 & 72.6$\pm$0.3 & 88.5 \\                            
      \textbf{CDAN + MCC}& \textbf{94.7}$\pm$0.2 & \textbf{98.6}$\pm$0.1 & \textbf{100.0}$\pm$0.0 & \textbf{95.0}$\pm$0.1 & 73.0$\pm$0.2 & \textbf{73.6}$\pm$0.3 & \textbf{89.2} \\
      \midrule
      AFN \cite{AFN2019}  & 88.8$\pm$0.5 & 98.4$\pm$0.3 & 99.8$\pm$0.1 & 87.7$\pm$0.6 & 69.8$\pm$0.4 & 69.7$\pm$0.4 & 85.7 \\
      AFN + MinEnt~\cite{Grandvalet2005Semi} &90.3$\pm$0.4 & \textbf{98.7}$\pm$0.2 & \textbf{100.0}$\pm$0.0 & 92.1$\pm$0.5 & 73.4$\pm$0.3 & 71.2$\pm$0.3 & 87.6 \\
      AFN + BSP~\cite{BSP2019} & 89.7$\pm$0.4 & 98.0$\pm$0.2 & 99.8$\pm$0.1 & 91.0$\pm$0.4 & 71.4$\pm$0.3 & 71.4$\pm$0.2 & 86.9 \\
      \textbf{AFN + MCC} &\textbf{95.4}$\pm$0.3 & 98.6$\pm$0.2 & \textbf{100.0}$\pm$0.0 & \textbf{96.0}$\pm$0.2 & \textbf{74.6}$\pm$0.3 & \textbf{75.2}$\pm$0.2 & \textbf{90.0} \\
      \bottomrule
   \end{tabulary}
\end{table}

\textbf{Ablation Study.} 
It is interesting to investigate the contribution of each part of the MCC loss: Class Correlation \textbf{(CC)}, Probability Rescaling \textbf{(PR)}, and Uncertainty Reweighting \textbf{(UR)}. Results in Table~\ref{table:ablation} justify that each part has its indispensable contribution. To enable ease of use, we seamlessly integrate these parts into a coherent loss and reduce the number of hyper-parameters.

\begin{table}[!htbp]
   \addtolength{\tabcolsep}{3pt}
   \centering
   \caption{Ablation study of MCC on Office-31 for UDA (ResNet-50).}
   \tiny
   \label{table:ablation}
   \scriptsize
   \begin{tabulary}{\linewidth}{l*{7}c}
      \toprule
      Method & A$\rightarrow$W & D$\rightarrow$W & W$\rightarrow$D & A$\rightarrow$D & D$\rightarrow$A & W$\rightarrow$A & Avg \\
      \midrule
      MCC (CC Only)   & 92.2  & 96.9 & \textbf{100.0} & 88.6  & 73.2  & 64.5  & 85.9 \\ 
      MCC (CC + PR)     & 93.1  & 98.5  & \textbf{100.0}  & 91.6 & 70.9  & 69.0 & 87.2 \\
      MCC (CC + PR + UR) & 93.7  & \textbf{98.6} & \textbf{100.0} & 93.2  & 72.1  & 73.7  & 88.4 \\ 
      \midrule
      \textbf{MCC (All)} & \textbf{95.5}  & \textbf{98.6}  & \textbf{100.0}  & \textbf{94.4}  & \textbf{72.9}  & \textbf{74.9} & \textbf{89.4} \\
      \bottomrule
   \end{tabulary}
\end{table}

Further, we analyze how the specially designed Uncertainty Reweighting (UR) mechanism works. Fig.~\ref{fig:weight} shows three typical examples as well as their weights and the confusion values before and after reweighting. 
The classifier prediction on the first image shows no obvious peak, while the one on the third image shows two obvious peaks on classes \texttt{calculator} and \texttt{phone}. The third image is more suitable for embodying class confusion. Naturally, its confusion value is higher than the first one, and our reweighting mechanism further highlights the suitable one. On the other hand, as the reweighting mechanism is defined with entropy, we recognize that it will improperly assign high weights to examples with highly confident predictions, including the wrong ones. As shown in the second image, its ground truth label is a \texttt{lamp}, but it is classified as a \texttt{bike}. In our method, the confusion value of such an example is so low that the influence of higher weight can be neglected. Therefore, our reweighting mechanism is effective and reliable.

\begin{figure}[htbp]
   \centering
      \includegraphics[width=0.9\textwidth]{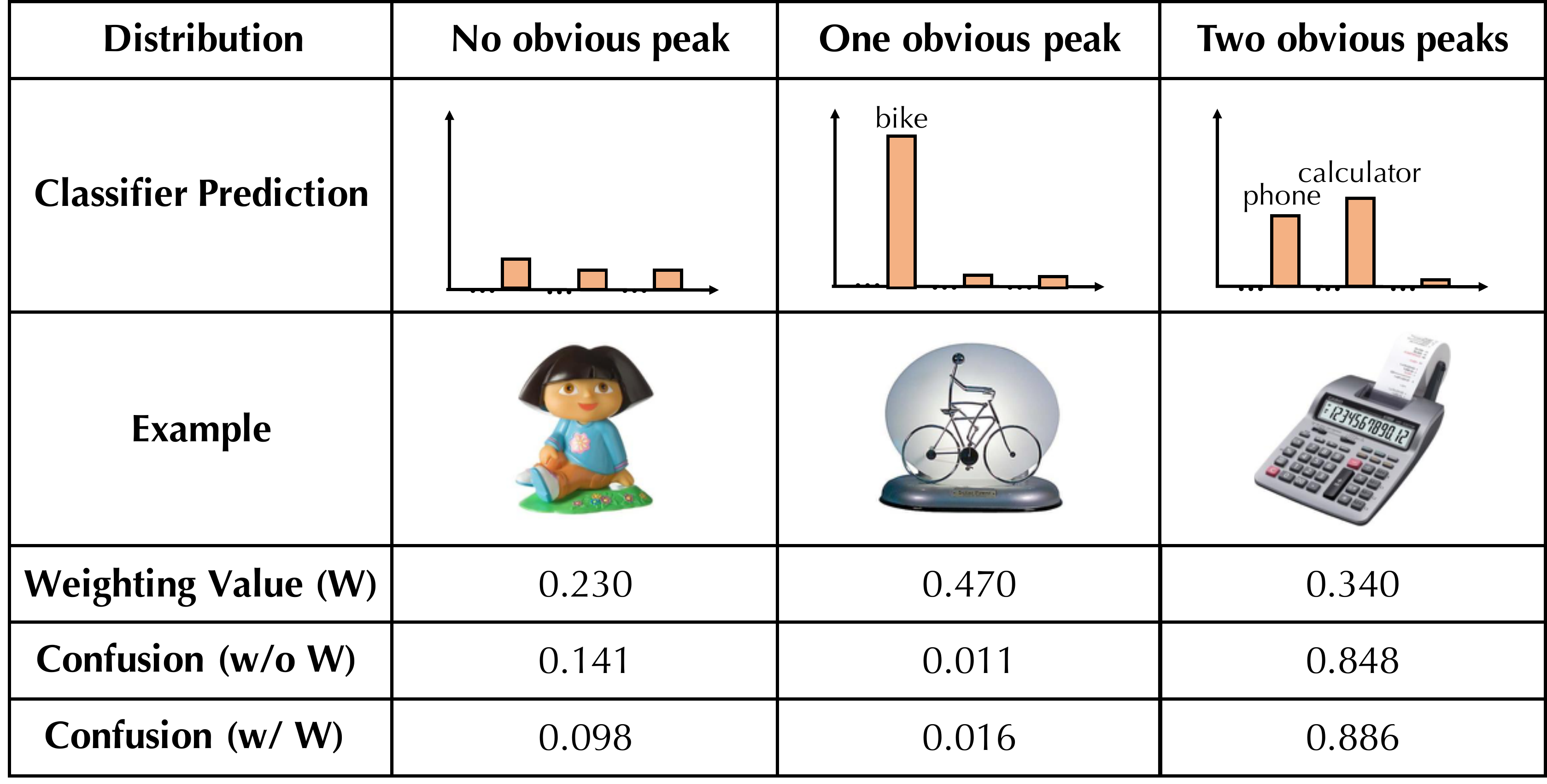}
      \caption{Three typical samples and the corresponding weights and confusion values.}
      \label{fig:weight}
\end{figure}

\textbf{Theoretical Insight.}
Ben-David \textit{et al.}~\cite{ben2010theory} derived the expected error $\mathcal{E}_\mathcal{T}(h)$ of a hypothesis $h$ on the target domain ${\mathcal{E}_\mathcal{T}}(h) \le {\mathcal{E}_\mathcal{S}}(h) + \frac{1}{2}{d_{\mathcal{H}\Delta \mathcal{H}}}(\mathcal{S},\mathcal{T}) + \epsilon_{\text{ideal}} $ by: \textbf{(a)} expected error of $h$ on the source domain, $\mathcal{E}_\mathcal{S}(h)$; \textbf{(b)} the A-distance ${d_{\mathcal{H}\Delta \mathcal{H}}}(\mathcal{S},\mathcal{T})$, a measure of domain discrepancy; and \textbf{(c)} the error $\epsilon_{\text{ideal}}$ of the ideal joint hypothesis $h^*$ on both source and target domains. As shown in Fig.~\ref{hyper}, MCC has the lowest A-distance~\cite{ben2010theory}, which is close to the oracle one (\textit{i.e.} supervised learning on both domains). In Fig. \ref{fig:converge-lambda}, the $\epsilon_{\text{ideal}}$ value of MCC is also lower than that of mainstream DA methods. Both imply better generalization.

\begin{figure}[htbp]
   \centering
   \subfigure[A-Distance~\cite{ben2010theory}]{
      \includegraphics[width=0.23\textwidth]{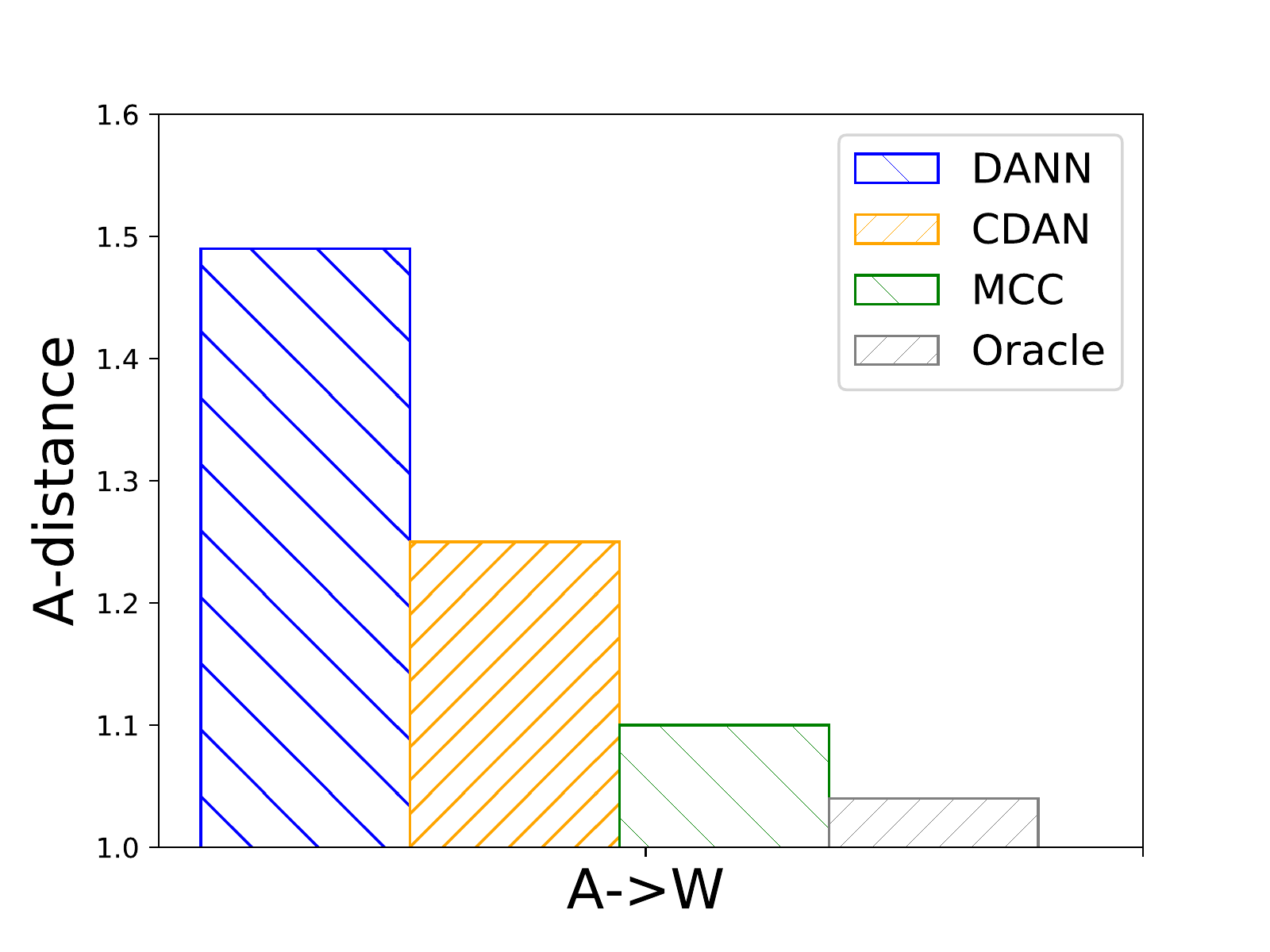}
   }
   \hfil
   \subfigure[MinEnt~\cite{Grandvalet2005Semi}]{
      \includegraphics[width=0.23\textwidth]{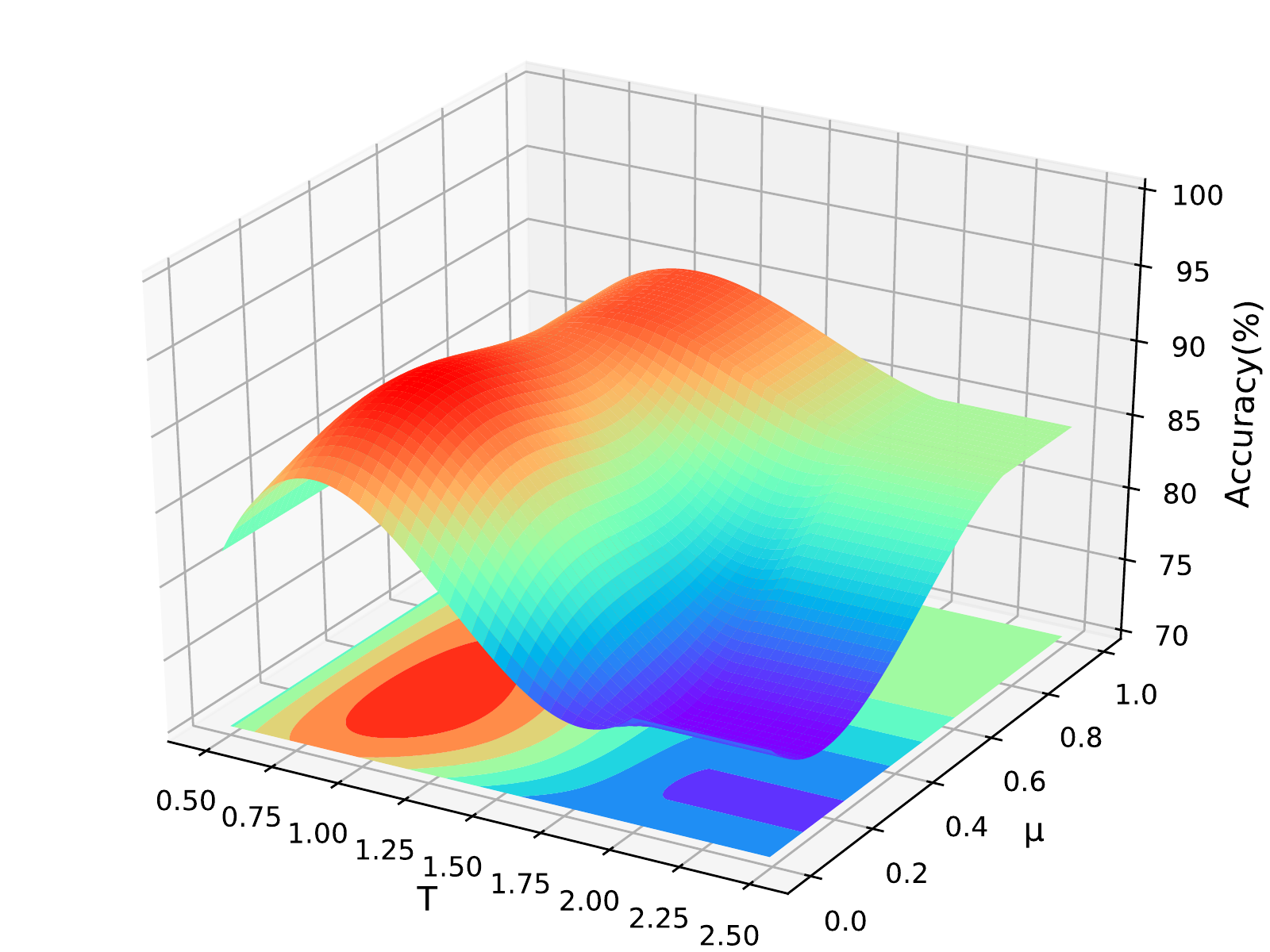}
   }
   \hfil
   \subfigure[MCC]{
      \includegraphics[width=0.23\textwidth]{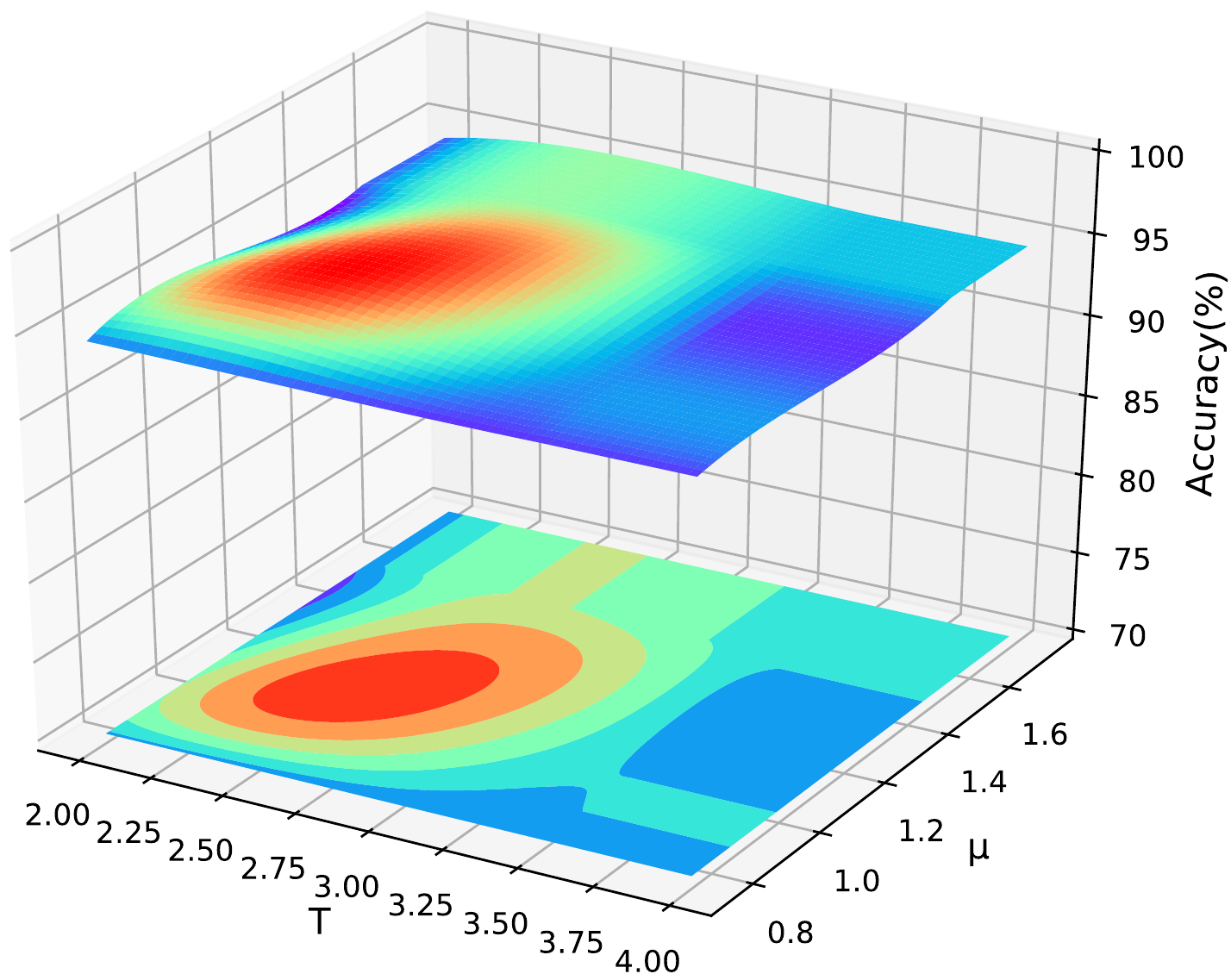}
   }
   \caption{(a): A-Distance of the last \textit{fc}-layer features of task $A \rightarrow W$ on Office-31 (UDA); (b)--(c): Hyper-parameter sensitivity of task $A \rightarrow W$ on Office-31 (UDA).}
   \label{hyper}
\end{figure}

\textbf{Parameter Sensitivity.}
Temperature factor $T$ and MCC coefficient $\mu$ are the two hyper-parameters of MCC and MinEnt \cite{Grandvalet2005Semi} when applying them standalone or with existing methods. We traverse hyper-parameters around their optimal values $[T^*, \mu^*]$, as shown in Fig. \ref{hyper}, MCC is much less sensitive to its hyper-parameters.

\textbf{Convergence Speed.}
We show the training curves throughout iterations in Fig. \ref{fig:converge-lambda}. Impressively, MCC takes only $1000$  iterations to reach an accuracy of $95\%$, while at this point the accuracies of CDAN and DANN are below $85\%$. 
When used as a regularizer for existing domain adaptation methods, MCC largely accelerates convergence. 
In general, MCC is $3 \times$ faster than the existing methods.

\begin{figure}[htbp]
   \centering
   \subfigure[]{
      \includegraphics[width=0.21\textwidth]{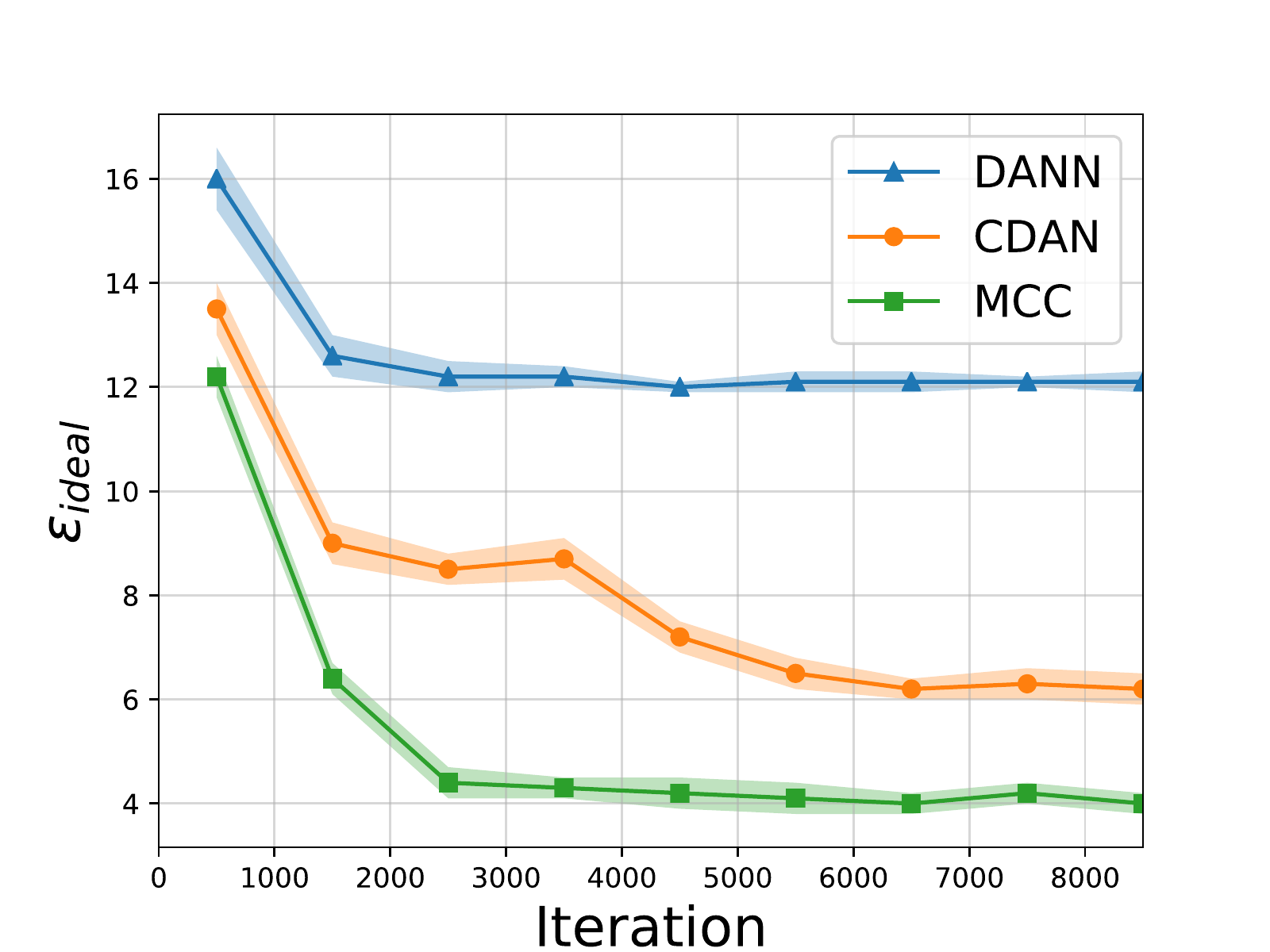}
   }
   \hfil
   \subfigure[]{
      \includegraphics[width=0.21\textwidth]{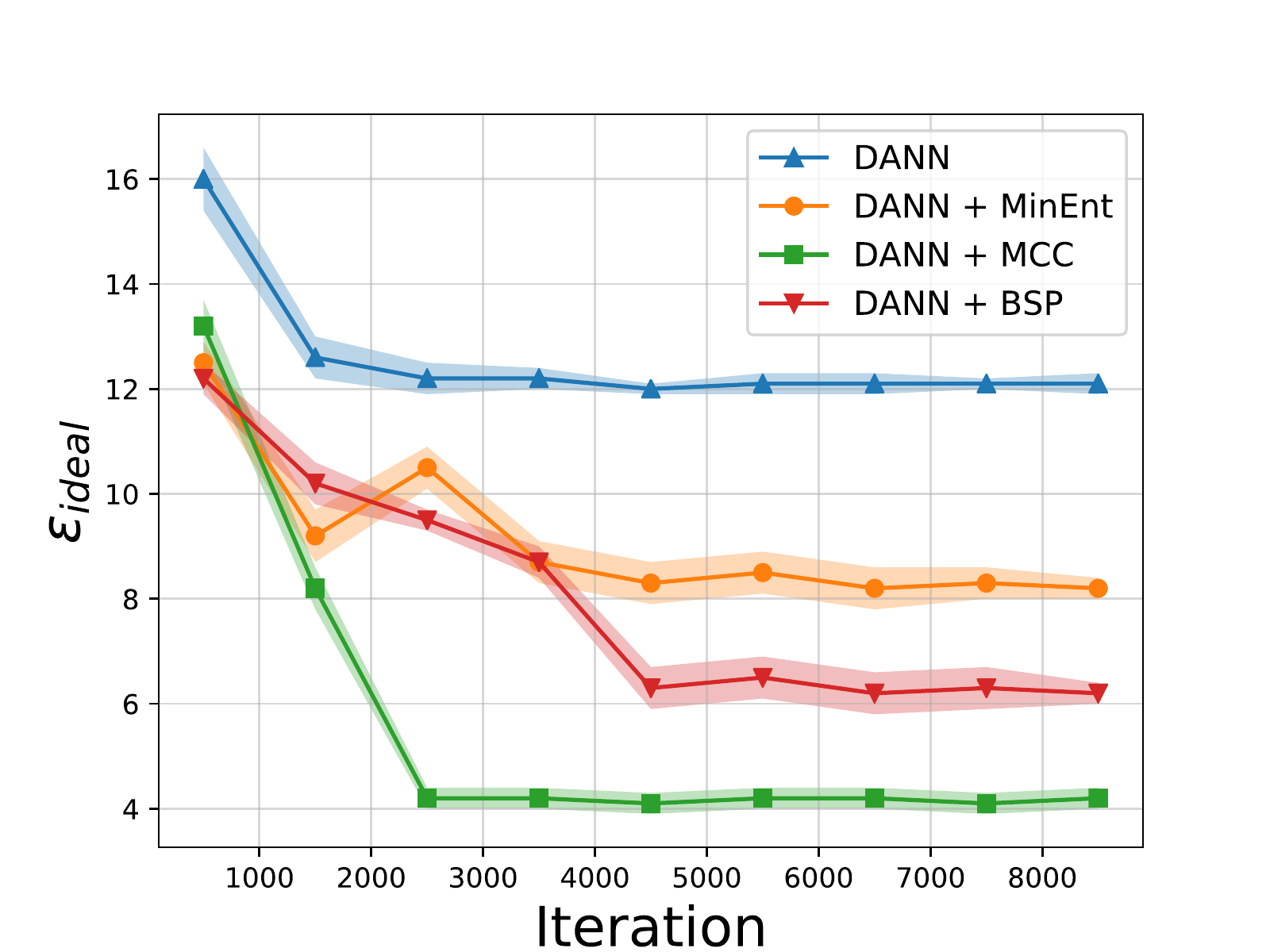}
   }
   \hfil
   \subfigure[]{
      \includegraphics[width=0.21\textwidth]{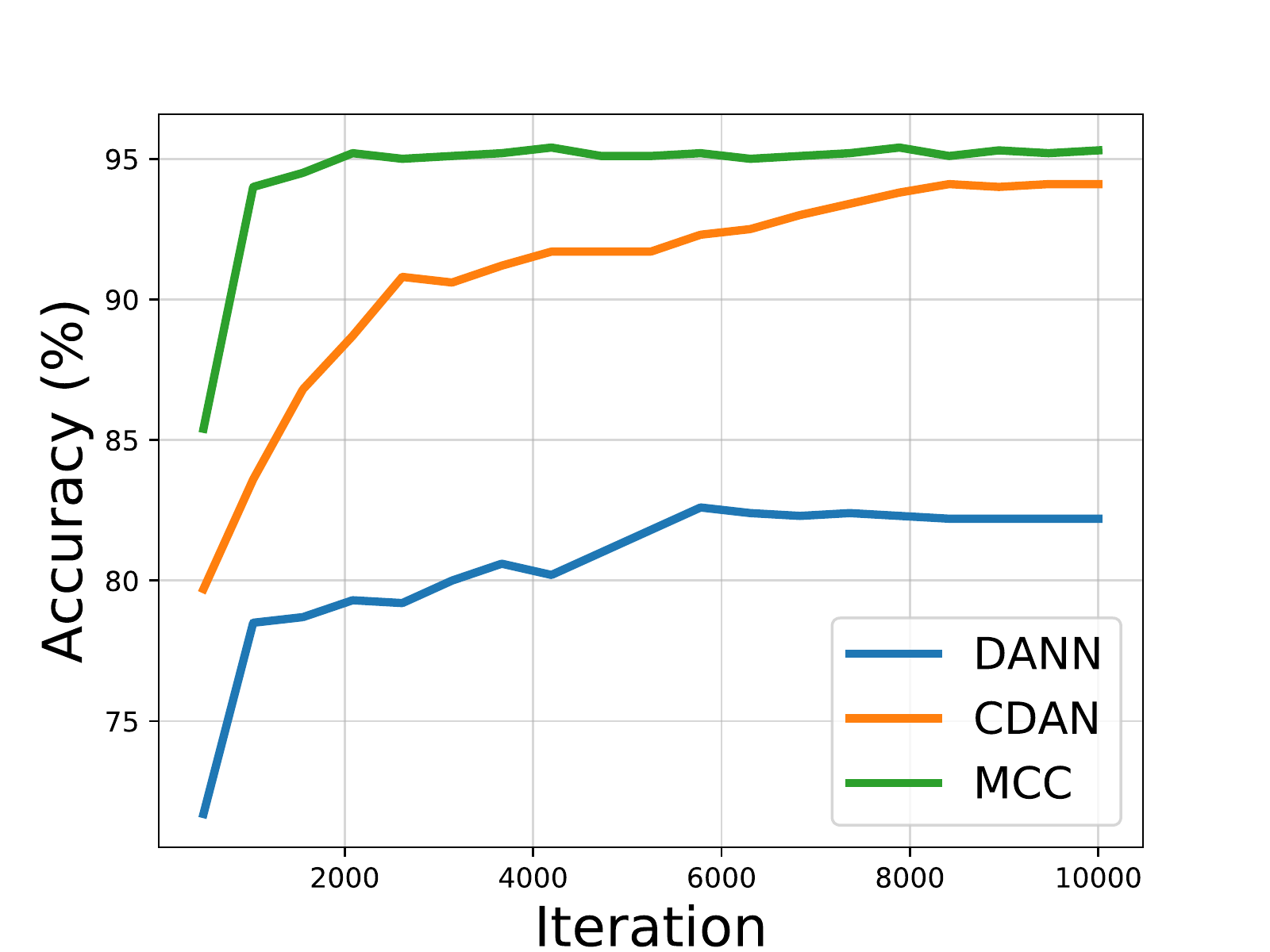}
      \label{MCC}
   }
   \hfil
   \subfigure[]{
      \includegraphics[width=0.21\textwidth]{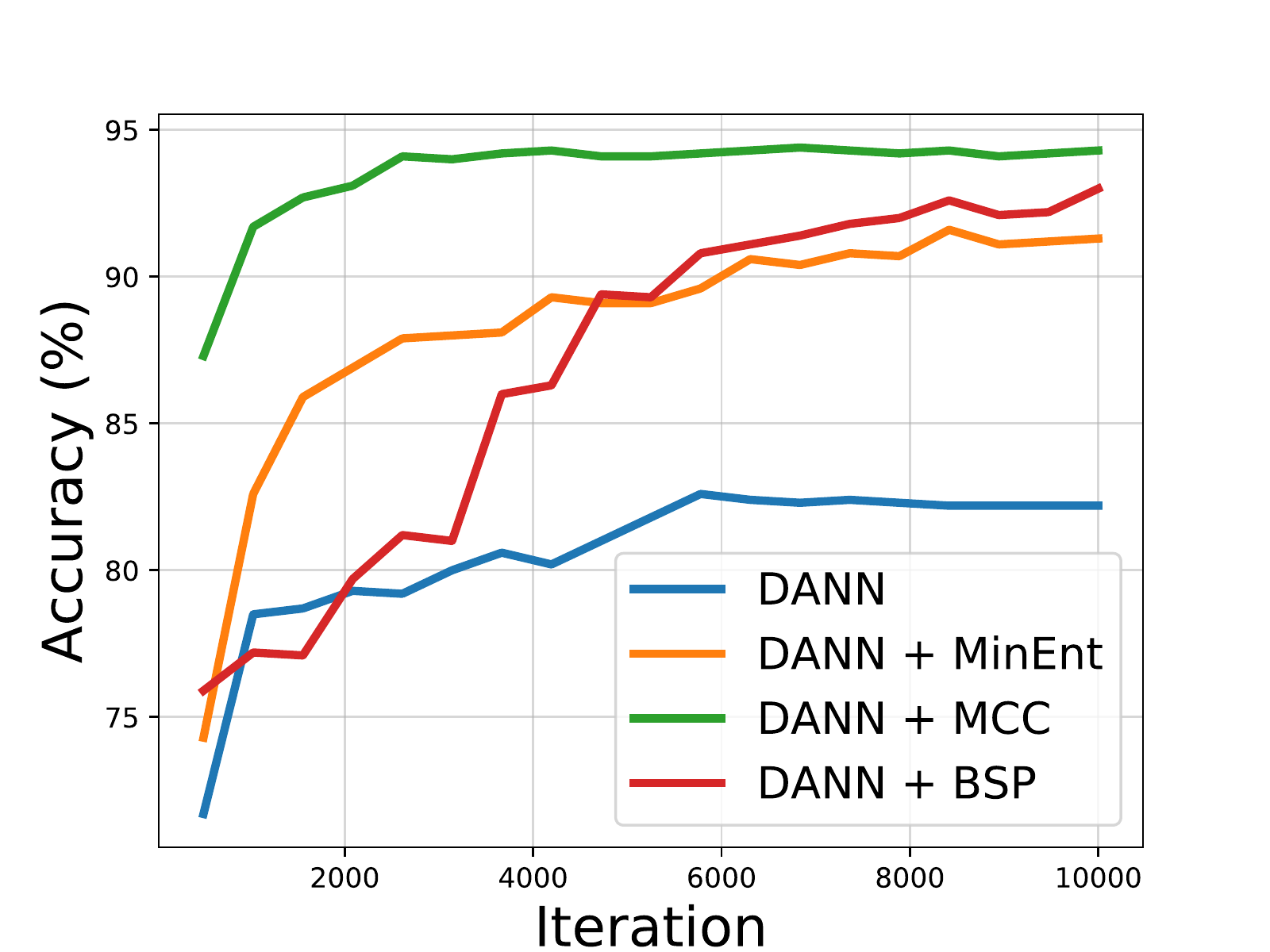}
      \label{MCC+DANN}
   }
   \caption{The $\epsilon_{\text{ideal}}$ error values (\%) and training curves throughout iterations.}
   \label{fig:converge-lambda}
\end{figure}

\section{Conclusion}
This paper studies a more practical paradigm, Versatile Domain Adaptation (VDA), where one method tackles many scenarios. We uncover that less class confusion implies more transferability, which is the key insight to enable VDA. Based on this, we propose a new loss function: Minimum Class Confusion (MCC). MCC can be applied as a versatile domain adaptation approach to a variety of DA scenarios. Extensive results justify that our method, without any modification, outperforms state-of-the-art scenario-specific domain adaptation methods with much faster convergence. Further, MCC can also be used as a general regularizer for existing DA methods, further improving accuracy and accelerating training.

\section*{Acknowledgments}
The work was supported by the Natural Science Foundation of China (61772299, 71690231), and China University S\&T Innovation Plan by Ministry of Education.

\clearpage

\bibliographystyle{splncs04}
\bibliography{egbib}
\end{document}